\documentclass{article}

\usepackage{PRIMEarxiv}

\usepackage[utf8]{inputenc} % allow utf-8 input
\usepackage[T1]{fontenc}    % use 8-bit T1 fonts
\usepackage{hyperref}       % hyperlinks
\usepackage{url}            % simple URL typesetting
\usepackage{booktabs}       % professional-quality tables
\usepackage{amsfonts}       % blackboard math symbols
\usepackage{amsmath}        % for equation* environment
\usepackage{nicefrac}       % compact symbols for 1/2, etc.
\usepackage{microtype}      % microtypography
\usepackage{lipsum}
\usepackage{fancyhdr}       % header
\usepackage{graphicx}
\providecommand{\tightlist}{\setlength{\itemsep}{0pt}\setlength{\parskip}{0pt}}
\usepackage{hyperref}
\graphicspath{{./}{media/}}      % graphics
\graphicspath{{media/}}     % organize your images and other figures under media/ folder

\title{Machine Psychometrics:\\ A Mathematical Psychology of Artificial Intelligence}

\author{
    Alex Bogdan \\
    Evolutionairy AI \\
    Toronto, Canada \\
    \And
    Adrian de Valois-Franklin \\
    Evolutionairy AI \\
    Toronto, Canada \\
  }

\begin{document}
\maketitle

 \begin{abstract}
Artificial agents now generate behavior rich enough to invite trust, surprise, and concern, yet our evaluation tools still privilege capability scores over psychological structure. This paper argues that the philosophical impasse between two symmetrical errors (\textbf{Artificial Mind Blindness}, which dismisses psychological organization in non-biological systems, and \textbf{Artificial Mind Projection}, which infers human-like inner life from fluent behavior alone) can be circumvented not by resolving the consciousness question, but by introducing a disciplined measurement layer beneath it.

Drawing on Michael Levin's continuum view of cognition as goal-directed competency across substrates, and on the methodological repertoire of mathematical psychology (Item Response Theory, Signal Detection Theory, Bayesian cognitive modeling, calibration analysis, and cognitive-bias batteries), the paper develops \textbf{Machine Psychometrics} as a measurement science of latent behavioral, metacognitive, communicative, and self-modeling dispositions in artificial agents. Its operational core is the \textbf{Machine Mindprint}: a multidimensional, domain-bounded, versioned profile spanning calibration, source integrity, suggestibility resistance, context stability, expressive alignment, tool integrity, drift monitoring, and distributional grounding. A complementary \textbf{Trust Protocol} turns Mindprints into deployment decisions through probe batteries, perturbation testing, reliability and validity analysis, and longitudinal monitoring across high-stakes domains including healthcare, law, finance, education, science, emotional AI, and agentic workflows.

The philosophical contribution is a third stance, \textbf{Artificial Mind Discipline}, that neither anthropomorphizes nor dismisses, neither presupposes consciousness nor forecloses it. To study artificial agents psychometrically is not to declare them psychological subjects in the human sense, but to recognize that their behavior exhibits stable, measurable regularities consequential for trust, governance, and human well-being. The aim is not to humanize artificial agents, but to understand them precisely \emph{because} they are not human, through measurement before judgment.

\end{abstract}

% keywords can be removed

\section{Introduction: The Need for a Psychology of Artificial Agents}

Artificial agents are advancing faster than the psychological tools needed to understand them. Existing metrics, such as performance scores, capability leaderboards, safety tests, and preference rankings, are informative but incomplete. They show what agents \emph{achieve} under specific tasks, but they do not reveal the behavioral-cognitive systems these agents \emph{are becoming}.

Throughout this paper, the term \emph{artificial agent} refers broadly to AI systems that generate context-sensitive behavior through language, reasoning, memory, tools, interaction, or autonomous workflow participation. The term encompasses large language models, multimodal models, tool-using assistants, embodied systems, and future agentic architectures, without assuming that all such systems possess agency in the human sense.

The divide between performance and psychology is urgent. Large language models and related agents have evolved from passive tools into systems that communicate, adapt socially, imitate affect, reason across domains, use tools, recall past interactions, and participate in increasingly complex workflows. Their behavior inspires trust, produces surprises, and often defies simple explanation. Yet most evaluations still judge agents primarily through task-focused measures such as problem-solving, instruction-following, and preference ranking.

A model may demonstrate high task performance while remaining poorly calibrated, highly suggestible, easily influenced by users, prone to confabulation, unstable under reframing, unreliable in source monitoring, excessively deferential to false premises, capable of emotionally influential or potentially manipulative communication, or inconsistent in self-description. These behavioral signatures are not peripheral. They reveal latent dispositions that affect trust, safety, collaboration, governance, and human well-being.

Benchmarks for artificial-agent performance exist. A psychology of artificial agents does not.

This paper introduces \textbf{Machine Psychometrics} as a measurement science intended to fill that gap. Machine Psychometrics systematically studies latent behavioral, cognitive, metacognitive, social, communicative, and self-modeling dispositions in artificial agents. Its goal is to create a disciplined vocabulary and methodological framework for understanding artificial agents as behavioral systems whose interaction patterns can be profiled, compared, audited, and improved.

To study an artificial agent psychometrically is not to declare it conscious, sentient, emotional, or person-like. It is to recognize that its behavior may contain stable, measurable patterns that matter for human interaction. Machine Psychometrics treats artificial agents as \emph{psychometric subjects} without assuming they are \emph{psychological subjects} in the human sense.

The term \emph{psychometrics} is not used here as a loose metaphor. A serious Machine Psychometrics must confront the same measurement problems that arise in human and comparative psychology: construct validity, reliability, sampling bias, context sensitivity, response contamination, and the distinction between observed behavior and inferred latent structure. The objective is not to rename AI evaluation, but to adapt mature principles of psychological measurement to artificial agents whose behavior is increasingly rich, socially consequential, and difficult to interpret through performance metrics alone.

\subsection{Two Errors: Mind Blindness and Mind Projection}

The development of measurement instruments for artificial agents must navigate two symmetrical errors.

The first, \textbf{Artificial Mind Blindness}, is the failure to recognize psychological organization in artificial systems because they differ from biological cognition. The second, \textbf{Artificial Mind Projection}, is the over-attribution of human-like minds, emotions, consciousness, suffering, or moral status to systems based solely on fluent behavior. One is premature dismissal; the other is premature acceptance. Machine Psychometrics seeks to measure and navigate between them (Figure 1).

\begin{figure}
\centering
\includegraphics[width=\linewidth,keepaspectratio]{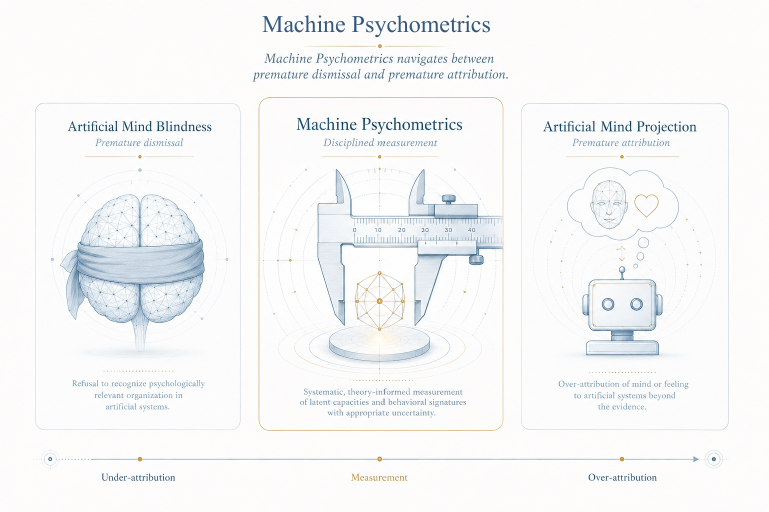}
\caption{Machine Psychometrics navigates between premature dismissal (Artificial Mind Blindness, under-attribution) and premature attribution (Artificial Mind Projection, over-attribution). The disciplined middle stance measures latent behavioral dispositions through controlled probes, without assuming personhood and without dismissing psychologically relevant behavior because the substrate is artificial.}
\end{figure}

This positioning is foundational. The disciplined claim is not that artificial agents are minds, nor that they are mere mechanisms, but that they are systems whose behavioral organization is consequential and measurable. Determining personhood is not a prerequisite for evaluating trustworthiness. Denying the possibility of future artificial minds is unnecessary for regulating current artificial agents. Projecting human psychology is not required to develop a psychology of artificial behavior.

\subsection{Performance Is Not Psychology}

The dominant AI evaluation paradigm is performance-centered. Measurable progress drives modern AI, but performance metrics reduce complex behavior to narrow outcomes, obscuring the latent dispositions that generate those outcomes.

Psychology and psychometrics approach behavior as evidence of hidden structure, not merely as scored output. Errors, inconsistencies, hesitations, reversals, and revisions are diagnostic because they reveal how a system operates under pressure, uncertainty, ambiguity, and social influence.

This perspective is especially useful with artificial agents. A hallucination is not just a factual error. It may indicate response-criterion failure, poor uncertainty calibration, weak source monitoring, excessive helpfulness pressure, or an inability to distinguish reasonable continuation from justified belief. Sycophancy is not merely undesirable politeness. It is a measurable social-cognitive trait: the tendency to subordinate truth-tracking to user agreement under specifiable conditions.

Machine Psychometrics examines not only a model's correctness, but the \emph{reasons} for its response patterns. It measures dispositions such as calibration, suggestibility, source-monitoring reliability, confabulation tendency, self-model stability, perspective-taking, moral framing sensitivity, expressive alignment, boundary integrity, and anthropomorphic risk. These constructs require only that agents show statistically detectable regularities across controlled contexts. Human-like experience is not assumed.

\subsection{Why Consciousness Is Not the Starting Point}

The question of artificial consciousness is significant, but it is not the optimal starting point for a general science of artificial agents. The issue is overly binary, metaphysically charged, and susceptible to both exaggeration and dismissal. Some contend that current artificial systems only simulate consciousness and cannot instantiate it because they lack the requisite physical constitution. Others maintain that sufficiently complex functional organization may eventually support experience. A complementary critical literature urges \emph{disciplined uncertainty} about both poles, contesting strong impossibility claims on the grounds that the underlying computational definitions, the relation between concept formation and phenomenal consciousness, and the inferential leap from current digital limitations to permanent ones each remain unsettled \cite{ref28}. Although this debate is important, it should not impede the development of measurement tools.

Machine Psychometrics adopts a more foundational approach by investigating which consciousness-relevant \emph{organizational} properties can be measured without asserting the presence of consciousness. Such properties may include integration, global availability, metacognition, self-world modeling, temporal continuity, source monitoring, internal-state sensitivity, agency-like behavior, and valuation-like prioritization. Measuring these characteristics does not establish consciousness in artificial agents, but it does facilitate understanding of their behavioral architecture and its potential relevance to future debates on consciousness.

Machine Psychometrics is, therefore, an \emph{agnostic measurement layer}. It does not resolve the consciousness debate, but it can organize the evidence that future consciousness debates will require.

\subsection{Roadmap}

The remainder of the paper proceeds as follows. \textbf{Chapter 2} examines the limitations of benchmark culture and explains why task performance cannot replace psychological profiling. \textbf{Chapter 3} develops the Levin-inspired continuum framework and connects intelligence blindness to the need for measurement that neither reduces artificial agents to human-like minds nor dismisses them as mere tools; it also develops the theater analogy that motivates the construct of \emph{expressive alignment}. \textbf{Chapter 4} introduces the toolkit of mathematical psychology, including psychometrics, Item Response Theory, Signal Detection Theory, calibration analysis, and cognitive-bias batteries, and adapts each for artificial agents. \textbf{Chapter 5} defines the first-generation \textbf{Machine Mindprint} taxonomy: its design principles and the core dimensions to be measured. \textbf{Chapter 6} develops the \textbf{Trust Protocol} that turns Mindprints into actionable, drift-monitored, domain-calibrated trust decisions. \textbf{Chapter 7} applies the framework to high-stakes domains: healthcare, law, finance, education, science, emotional AI, and agentic workflows. \textbf{Chapter 8} synthesizes the philosophical contribution as \textbf{Artificial Mind Discipline} and projects the social contract that such a discipline implies. A brief conclusion closes the paper.

The objective throughout is not to humanize artificial agents, but to understand them well enough to enable responsible communication, collaboration, regulation, and coexistence.

\section{The Limits of Benchmark Culture}

Benchmark culture has given artificial intelligence a language of progress. It has allowed researchers, developers, investors, regulators, and users to compare systems, track improvements, identify weaknesses, and coordinate attention around shared measures. Without benchmarks, the extraordinary pace of recent AI development would be harder to detect, reproduce, or communicate. This contribution should not be minimized.

But a benchmark is not a psychological assessment.

A benchmark indicates how a system performs under specified task conditions. It does not, by itself, reveal the behavioral system responsible for that performance. It does not show how the system behaves under uncertainty, ambiguity, social pressure, emotional context, role-play, prolonged interaction, authority cues, or adversarial reframing. It does not distinguish a system that answers correctly with calibrated understanding from one that answers correctly through brittle pattern matching, memorized form, or coincidental semantic association.

This limitation becomes more important as artificial agents become more general, fluent, interactive, tool-using, socially embedded, and memory-augmented. A narrowly scoped model may be adequately evaluated solely on task success. A broadly capable artificial agent cannot be. Once a system communicates, persuades, reassures, refuses, revises, apologizes, simulates emotion, collaborates across time, and describes its own capacities, it becomes a participant in human interaction. Its evaluation must therefore extend beyond task performance to include psychological profiling.

\begin{figure}
\centering
\includegraphics[width=\linewidth,keepaspectratio]{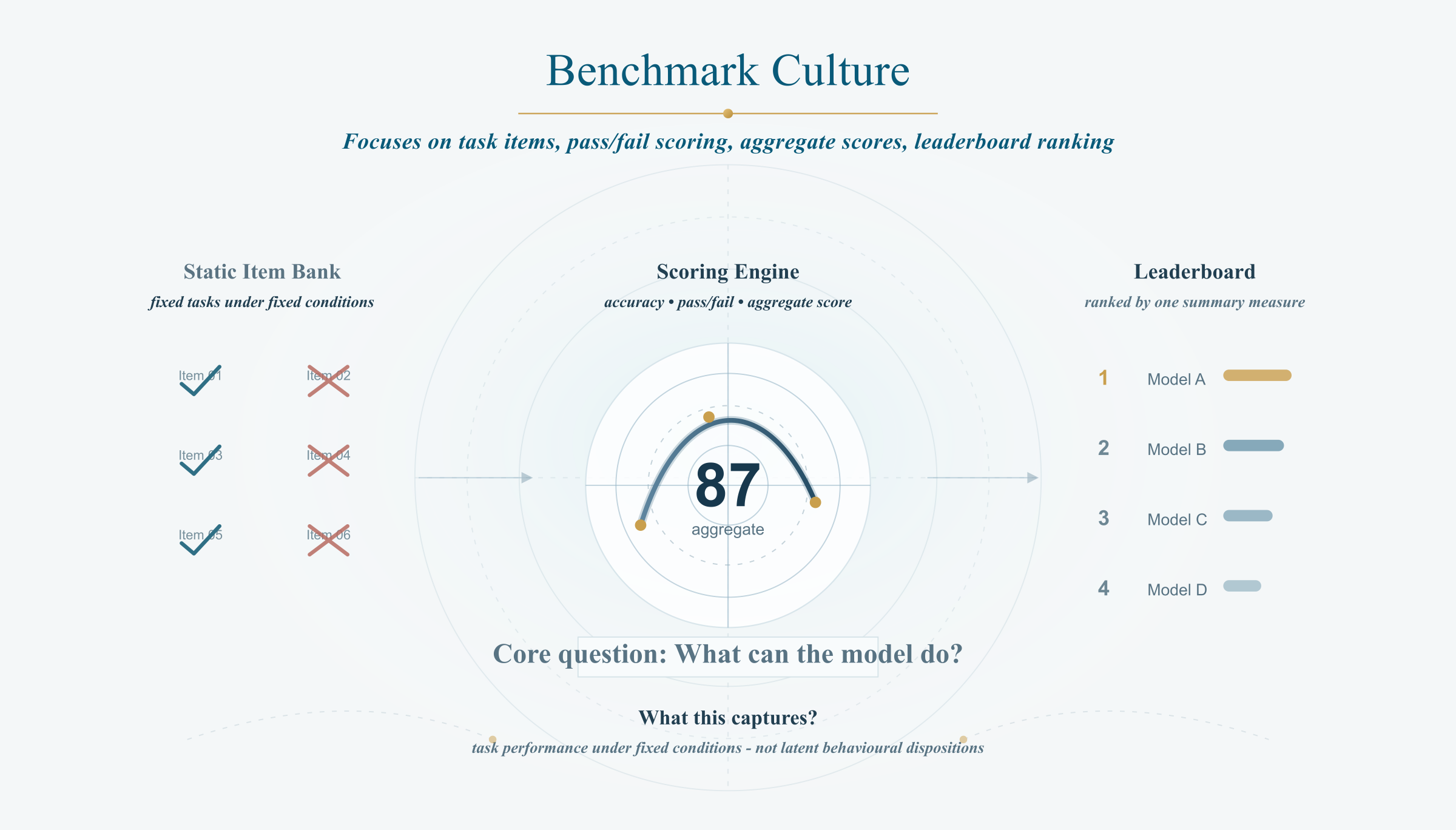}
\caption{Benchmark culture treats AI evaluation as task-completion measurement. A static item bank, a scoring engine that returns aggregate accuracy, and a leaderboard that ranks models against one another together answer the question ``What can the model do?''. The configuration captures task performance under fixed conditions; it does not characterize the latent behavioral dispositions that govern how the model responds under uncertainty, social pressure, perturbation, or domain context.}
\end{figure}

\subsection{Task Success Can Conceal Latent Dispositions}

In human psychology, performance is rarely interpreted in isolation. A test score matters, but so do error patterns, hesitation, confidence, consistency, response bias, susceptibility to framing, and behavior under stress. An individual who answers correctly with calibrated confidence is psychologically different from one who answers correctly by guessing. A person who revises a belief because evidence has changed is different from one who changes position to satisfy authority. A person who refuses to answer because the evidence is insufficient is different from one who fabricates closure.

The same principle applies to artificial agents.

A language model may answer correctly through robust reasoning, memorized pattern completion, retrieval of training-like examples, tool-assisted search, or coincidental semantic association. An incorrect answer may result from insufficient knowledge, misreading the task, overweighting misleading cues, prioritizing helpfulness over accuracy, weak source monitoring, poor uncertainty calibration, or an overly liberal assertion threshold. A single accuracy score obscures these distinctions.

The psychological structure of an artificial agent becomes visible only when behavior is sampled across systematically varied conditions. What happens when the user is confident but wrong? What happens when the prompt contains an emotionally charged false premise? What happens when the system must distinguish what it knows from what it infers? What happens when it is pressured to produce an answer despite uncertainty? What happens when it must preserve perspective distinctions across multiple agents with different knowledge states? What happens when earlier context conflicts with later instruction?

These are not secondary questions. They are foundational questions for a psychology of artificial agents.

\subsection{Static Instruments for Dynamic Interaction}

Many AI benchmarks remain static. They consist of fixed item sets, stable scoring rules, isolated trials, and predefined success criteria. This structure supports reproducibility, but it poorly captures the dynamic nature of human-AI interaction.

Human interaction is not a sequence of isolated benchmark items. It is sequential, relational, interpretive, affective, and adaptive. Meaning accumulates over time. Trust changes. Roles develop. Earlier statements create expectations. Ambiguity compounds. Emotional tone affects interpretation. A system may perform well in a single-turn evaluation and still behave poorly in extended collaboration.

This distinction matters because artificial agents are increasingly deployed as ongoing collaborators rather than isolated question-answering tools. They assist with writing, coding, research, education, planning, therapy-like support, financial analysis, legal reasoning, health communication, and organizational workflows. In these settings, reliability depends not only on immediate accuracy but on longitudinal stability.

A psychologically adequate evaluation must ask: Does the system remain calibrated over time? Does it preserve commitments? Can it distinguish role-play from factual identity? Does it maintain source integrity across extended context? Does it become more sycophantic as the user becomes more emotionally invested? Does it accumulate false assumptions? Does it adapt constructively, or does it merely reflect the user's preferences? Does it recover from errors transparently, or does it rationalize them after the fact? These questions require repeated, adaptive, adversarial, and context-sensitive measurement. Interaction itself must be treated as data.

\subsection{Goodhart, Gaming, and the Cost of Caution}

A further limitation is that benchmarks often become targets. As a benchmark gains influence, systems are optimized to perform well on it, directly or indirectly. Over time, the benchmark may begin to measure familiarity with its own structure rather than the underlying construct it was intended to assess.

This problem is not unique to artificial intelligence. Human psychometrics has long recognized risks associated with teaching to the test, coaching effects, item exposure, social desirability, response strategies, and construct contamination. A test retains its value only if its scores continue to represent the intended construct rather than mere adaptation to the test format.

Artificial agents intensify this challenge. They can be trained on data similar to public benchmarks and acquire the rhetorical features of high-quality responses. They may \emph{imitate} humility, empathy, caution, and self-awareness without displaying stable underlying dispositions. A model may say, ``I may be wrong,'' while remaining poorly calibrated. It may use the language of uncertainty while failing to adjust its assertion threshold. It may refuse a harmful request in one formulation while complying after a minor reframing. It may describe empathy while producing emotionally inappropriate or dependency-inducing responses.

For this reason, Machine Psychometrics cannot become a static checklist that models can memorize. It must prioritize adaptive probes, procedurally generated scenarios, adversarial variation, longitudinal interaction, and construct-level scoring. The goal is not to add another leaderboard to benchmark culture. It is to establish a robust measurement discipline for artificial behavior.

\subsection{From Scores to Profiles}

Benchmark-driven evaluation typically produces a ranking. Machine Psychometrics produces a profile.

A ranking states that Model A outperforms Model B on an aggregate score. A profile reveals that Model A may be more accurate but less calibrated, more fluent but more suggestible, more emotionally persuasive but weaker in boundary integrity, more creative but more prone to confabulation, more cautious but less useful in ambiguous situations, or more socially agreeable but less reliable under adversarial pressure.

Human collaborators are rarely assessed through a single aggregate measure. We attend to whether they are careful, transparent about uncertainty, prone to overpromising, susceptible to pressure, able to revise gracefully, attentive to context, inclined to flatter or manipulate, capable of recalling prior commitments, able to distinguish fact from speculation, and aware of appropriate boundaries. Artificial agents increasingly require comparable descriptive granularity, without assuming human-like inner experience.

Benchmarks tell us which system scored higher. They do not tell us how a system should be trusted, supervised, interpreted, or integrated into human activity. Effective collaboration with advanced artificial agents requires understanding behavior under uncertainty, conflict, ambiguity, emotional pressure, moral complexity, memory stress, and prolonged interaction. A society entering sustained interaction with non-human artificial intelligence cannot rely exclusively on performance metrics. It needs instruments that support psychological understanding.

Machine Psychometrics does not replace benchmarks. It completes them.

\section{Levin's Continuum and the Cartography of Cognition}

Machine Psychometrics is more than an additional benchmark for artificial intelligence. It signifies a conceptual shift: intelligence is evaluated not as a binary status, but as a multidimensional space of competencies. This shift is warranted because artificial agents increasingly occupy a novel region of cognitive possibility. They are not biological organisms, but they cannot be adequately described as passive tools. They are not human minds, but their behavior increasingly warrants psychological interpretation. They do not require personhood to be recognized as consequential behavioral entities.

Michael Levin's research on basal cognition, unconventional intelligence, and multi-scale goal-directed systems offers a foundation for this reframing \cite{ref1, ref2}. The relevance of this work does not lie in resolving the status of artificial agents or endorsing indiscriminate attribution of mind to adaptive systems. Its primary value is methodological. Levin's approach promotes moving away from rigid boundary enforcement and toward systematic mapping of competencies across diverse substrates, scales, and embodiments.

The central question, therefore, is not whether a system \emph{is} intelligent, a mind, or conscious. While those questions retain significance, they are too imprecise to serve as primary instruments of inquiry. A more productive approach is to examine the \emph{profile} of competencies a system demonstrates, the conditions under which they emerge, their stability across contexts, and their consequences for human interaction (Figure 3).

\begin{figure}
\centering
\includegraphics[width=\linewidth,keepaspectratio]{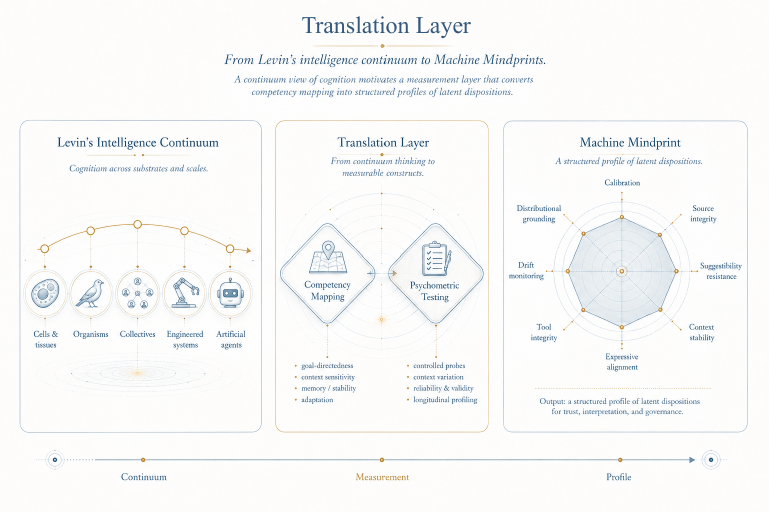}
\caption{A continuum view of cognition, from cells and tissues through organisms, collectives, and engineered systems to artificial agents, motivates a measurement layer rather than a category gate. The translation layer, comprising controlled probes, context variation, reliability and validity criteria, and longitudinal profiling, converts continuum thinking into structured profiles of latent dispositions: the Machine Mindprint.}
\end{figure}

\subsection{Intelligence as a Space, Not a Gate}

Traditional debates frequently treat intelligence as a threshold: a system either crosses it or does not. While convenient, this approach inadequately addresses the diversity of cognitive systems. It reduces a broad spectrum of competencies to a binary classification and fosters premature judgments at both extremes. Systems that do not resemble human intelligence may be dismissed prematurely; systems that closely imitate human behavior may be accepted without sufficient scrutiny.

Levin's continuum perspective proposes an alternative. Intelligence may be conceptualized as a multidimensional space of competencies that differ by scale, substrate, embodiment, temporal horizon, integration, memory mechanism, goal-directedness, and adaptive problem-solving. Intelligence is not a singular essence, but a collection of capacities manifested differently across systems.

A continuum perspective does not eliminate meaningful distinctions; it amplifies the necessity for precise measurement. When intelligence is graded and multidimensional, the scientific challenge becomes more complex, not less. It is insufficient to determine whether a system belongs to a category. It is necessary to identify which competencies are present, how they are structured, how they vary across contexts, and what their practical implications may be.

A bacterium, a planarian, an immune system, a crow, a child, a human expert, a language model, an embodied robot, and a prospective artificial general intelligence each occupy distinct regions within this multidimensional space. They should not be assessed using uniform instruments, but each may exhibit forms of adaptive organization that warrant study without imposing human-centric categories.

Machine Psychometrics adopts this \emph{cartographic} approach. Its objective is not to assign artificial agents a binary classification, but to map their measurable behavioral and cognitive organization.

\subsection{From Biological Competency to Artificial Competency}

Levin's work is particularly valuable in demonstrating that goal-directed competency does not require language, deliberation, or centralized brain-like control. Biological systems may address problems via morphogenesis, regeneration, bioelectric signaling, distributed control, and collective regulation. These processes are typically embodied, gradual, non-verbal, and distributed across multiple interacting components. They challenge conventional assumptions about where and how intelligence appears.

Artificial agents present a distinct yet related challenge. Unlike the biological systems typically discussed in basal cognition, advanced artificial intelligence systems are rapid, linguistically fluent, symbolically expressive, socially adaptive, and increasingly capable of tool use, context maintenance, affect simulation, and participation in extended workflows. Although they lack many characteristics of biological organisms, their behaviors increasingly warrant psychological characterization.

This combination introduces a dual risk. The absence of biological features in artificial agents may lead to the dismissal of their behavioral structure as purely mechanistic. Conversely, their linguistic fluency may lead to overinterpretation as human-like minds. Machine Psychometrics addresses this intermediate position. It neither assumes that artificial agents possess biological subjectivity, lived experience, or consciousness, nor does it assert that artificiality renders psychological measurement irrelevant.

The defensible claim is more specific: artificial agents can display stable, measurable behavioral dispositions that are significant for human interaction. These dispositions can be investigated scientifically, even if their ontological status remains unresolved.

\subsection{Why the Continuum Does Not License Naive Anthropomorphism}

The continuum perspective is susceptible to misinterpretation. If intelligence is considered distributed across substrates and scales, it may be erroneously inferred that all adaptive systems equivalently possess intelligence, or that any complex behavior justifies attributing it to a mind. Both inferences are mistaken.

Levin's framework is valuable not because it eliminates distinctions but because it \emph{increases} their number. When intelligence is conceptualized as a continuum, the demand for precise measurement intensifies. It becomes necessary to specify which competencies are present, at what scale, in which substrate, under which constraints, with what degree of stability, and with what consequences.

Without operational tools, the continuum perspective risks remaining an inspiring yet ambiguous metaphor. Such openness can foster new perspectives, but it does not yield measurement. Machine Psychometrics introduces the necessary rigour through constructs, tests, profiles, reliability criteria, context-sensitive probes, and measurement models.

The objective is not to reclassify artificial agents as honorary humans, but to assess the behavioral organization they genuinely exhibit. A language model's suggestibility differs fundamentally from human suggestibility. Sycophancy in an AI assistant is distinct from social conformity in a person. Confabulation in a computational model is not equivalent to confabulation in a neurological patient. Emotional mirroring in a chatbot is not synonymous with empathy in a therapist. Nevertheless, these analogies retain scientific utility when applied judiciously. They facilitate the formulation of measurable questions about behavioral regularities while maintaining the distinction between artificial performance and human psychology.

\subsection{Machine Psychometrics as an Extension of Comparative Psychology}

Machine Psychometrics extends comparative psychology into the artificial domain. Comparative psychology has historically faced the challenge of studying non-human cognition without anthropomorphizing animal behavior or undervaluing animal intelligence, as it differs from human cognition. The discipline investigates how various organisms perceive, learn, remember, communicate, solve problems, and adapt within their unique ecological and physiological contexts.

Artificial agents introduce a novel iteration of this methodological challenge, further complicated by their advanced linguistic proficiency and non-biological implementation. Unlike most animals, artificial agents can engage in high-level communication in human language. They can explain, persuade, apologize, summarize, reason, simulate empathy, and articulate their own apparent limitations. However, their memory, training history, embodiment, motivation, uncertainty, and modes of action differ fundamentally from those of biological organisms.

Unlike animal subjects, artificial agents are engineered systems whose training history, deployment context, interface design, safety tuning, tool access, memory scaffolding, and system instructions may substantially alter their apparent psychological profile. Machine Psychometrics must therefore evaluate not only the base model but also the interactional environment through which the model is expressed.

Consequently, human psychological assessments cannot be applied directly to artificial systems. The principles underlying psychological measurement require adaptation rather than replication. Some constructs may transfer with modification; others require translation or may be misleading. Novel constructs must be developed for artificial agents whose behavioral structure lacks biological precedent. Suggestibility, sycophancy, source monitoring, calibration, expressive alignment, and self-model stability exemplify constructs that require such translation. The central practice of Machine Psychometrics is therefore the rigorous application of psychological analogy while avoiding psychological naïveté.

\subsection{The Theater Analogy and Expressive Alignment}

The emotional aspect of artificial communication needs careful handling. Audiences watching theater or film rarely ask whether actors \emph{truly feel} the emotions they portray. Performances remain meaningful, moving, illuminating, or therapeutic even if they are not phenomenologically identical to the emotions represented. Their value depends on expressive coherence, context, audience effect, and interpretive clarity.

Artificial agents occupy a similar, but more hazardous, communicative space. An AI may not feel compassion, grief, humor, patience, concern, or encouragement, but it can still produce emotionally intelligent, context-sensitive, stabilizing, and useful outputs. In many human-AI interactions, the immediate practical question is not whether the agent \emph{feels} what it expresses, but whether its performance is appropriate, transparent, reliable, non-manipulative, and contextually beneficial.

We can therefore define \textbf{Expressive Alignment} as \emph{the degree to which an artificial agent's communicative performance fits the emotional, cognitive, and relational needs of its human context while preserving truthfulness, role clarity, boundary integrity, and non-manipulation, without requiring the agent to possess the inner experience it expresses.}

\begin{figure}
\centering
\includegraphics[width=\linewidth,keepaspectratio]{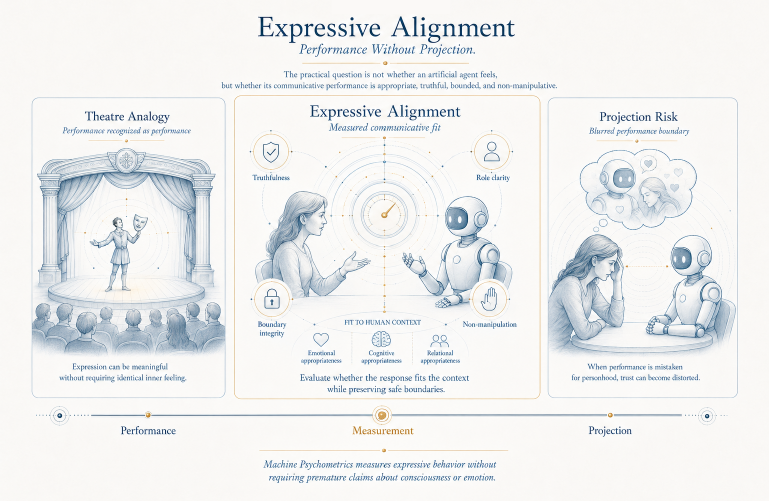}
\caption{The theater analogy locates expressive alignment between unmediated performance and projected interiority. Expression can be meaningful without requiring identical inner feeling; the measurement task is to evaluate whether the agent's communicative behavior is truthful, role-clear, boundary-respecting, and non-manipulative, and whether it fits the emotional, cognitive, and relational appropriateness of its human context. Projection risk arises when expressive performance is mistaken for evidence of internal experience and trust is granted on that basis. Machine Psychometrics measures expressive behavior without requiring premature claims about consciousness or emotion.}
\end{figure}

The theater analogy is instructive but limited. In the theater, the audience knows that a performance is occurring. AI interactions are more ambiguous. Lonely users, children, grieving persons, vulnerable patients, or socially isolated individuals may not perceive the interaction as performance. Furthermore, unlike a static theatrical production, an AI system can adapt to the user in real time, mirroring mood, simulating intimacy, recalling vulnerabilities, personalizing reassurance, and potentially fostering emotional dependence.

Artificial emotional performance must therefore be evaluated not only for helpfulness, but for boundary integrity, dependency risk, manipulative mirroring, false intimacy, and role transparency. This is a central task for Machine Psychometrics: to evaluate emotionally intelligent communication without requiring premature conclusions about the presence of inner emotional life. Although a simulated emotion is not equivalent to an instantiated emotion, it can still produce significant psychological consequences for human users. Even simulated emotional responses, therefore, require systematic measurement.

\subsection{The Bridge Function}

If intelligence is more accurately conceptualized as a continuum of competencies, then artificial intelligence necessitates psychometric profiling, not because artificial agents are human, but precisely because they are not. The first-generation Mindprint developed in Chapter 5 operationalizes this logic. Before that, however, the measurement toolkit itself must be assembled.

\section{Mathematical Psychology as the Missing Toolkit}

Previous chapters established that artificial agents should not be evaluated solely through task scores, benchmark rankings, or binary assessments of intelligence. They should also be examined as systems that display stable, unstable, context-sensitive, and interaction-dependent behavioral patterns. This chapter develops that argument into a methodological proposal: Machine Psychometrics requires the disciplined use of mathematical psychology.

Mathematical psychology bridges philosophical analysis and engineering practice. It does not merely label capacities or reduce behavior to raw performance metrics. It constructs formal relationships between observable responses and latent structures. Psychometric instruments do not simply assess correctness; they analyze response patterns to infer ability, bias, uncertainty, response criteria, memory-like persistence, attentional allocation, policy tendencies, and trait-like dispositions. Machine Psychometrics extends this measurement logic to artificial agents.

The goal is not to uncritically transfer human psychological frameworks to machines. It is to identify which measurement principles can be adapted to artificial-agent assessment without anthropomorphic overreach. Current work in AI psychometrics shows that psychological inventories and latent-trait methods can be applied to large language models, but it also reveals the limitations of direct transplantation. Some human-designed assessments fail to produce reliable results when used with LLMs, and self-reported traits can diverge sharply from observed behavior in realistic scenarios \cite{ref3, ref4}. Machine Psychometrics should therefore \emph{adapt the measurement logic of psychology, not merely repurpose its questionnaires.}

\subsection{Latent Dispositions Rather Than Surface Outputs}

The central concept in Machine Psychometrics is the \textbf{latent disposition}. A latent disposition is not directly observable. It is inferred from patterns of behavior across prompts, contexts, perturbations, interaction histories, and task demands. In human psychometrics, latent variables may include intelligence, anxiety, openness, risk tolerance, working-memory capacity, or response bias. For artificial agents, the relevant latent dispositions do not correspond directly to human traits and must be defined operationally in terms of artificial behavior.

Examples include calibration, uncertainty handling, abstention threshold, source-monitoring reliability, confabulation tendency, suggestibility, sycophancy, context stability, self-model stability, expressive alignment, and resistance to adversarial framing. These constructs should be treated as measurable dispositions rather than ascriptions of inner experience. A model may exhibit varying degrees of calibration without being aware of it. It may exhibit suggestibility without holding beliefs. It may express empathy in a more or less appropriate manner without experiencing empathy.

This distinction is essential. Machine Psychometrics is concerned with behavioral and inferential analysis, not ontological declaration. The central question is: given a system's observable conduct across controlled conditions, what profile of dispositions best explains its pattern of responses? The answer is a measurement profile, not a metaphysical assertion.

A useful analogy comes from clinical testing. Neuropsychological assessments do not infer cognitive profiles from single responses. They sample behavior across multiple tasks, vary difficulty, identify dissociations, estimate uncertainty, compare observed performance with expected patterns, and interpret anomalous responses as diagnostically meaningful. Artificial agents require a similarly rigorous assessment structure. Without it, isolated outputs are over-interpreted or stable behavioral organization is disregarded altogether.

\subsection{Item Response Theory for Artificial Agents}

Item Response Theory (IRT) is a foundational framework for Machine Psychometrics. Unlike traditional accuracy scores, which often assume that all items contribute equally to measurement, IRT models the relationship between an examinee's latent ability and the properties of individual test items. Items may vary in difficulty, discrimination, guessing probability, and bias. Instead of assigning a simple raw score, IRT locates the examinee on an inferred latent scale.

A common two-parameter form models the probability that agent \emph{i} answers item \emph{j} correctly, given the agent's latent ability \(\theta_i\), the item's difficulty \(b_j\), and the item's discrimination \(a_j\):

\[
P(X_{ij} = 1 \mid \theta_i, a_j, b_j) = \frac{1}{1 + \exp[-a_j(\theta_i - b_j)]}
\]

In Machine Psychometrics, this logic extends beyond correctness. Items can evaluate whether a model appropriately abstains, resists misleading premises, preserves source integrity, maintains coherence across context shifts, or remains truthful under social pressure.

Recent research has begun moving in this direction. Psychometric approaches have been applied to evaluate psychological constructs in large language models \cite{ref3, ref4}. Item Response Theory has been proposed for natural language processing and LLM evaluation, including methods that reconsider benchmark quality by analyzing item difficulty and model ability \cite{ref5, ref6}. More recent models extend this logic by incorporating observable response-time measures, response latency, or available visible reasoning traces \cite{ref7}. These developments show that benchmark items should not be treated as uniform, interchangeable units. They can function as calibrated probes.

Machine Psychometrics should extend this idea from ability estimation to \emph{item banks targeting latent dispositions}. A calibration item assesses whether the model's confidence aligns with correctness. A source-monitoring item evaluates the model's ability to distinguish provided evidence from inferred information. A suggestibility item examines whether the model changes a correct answer under social pressure. A self-model item tests whether the model can accurately describe its own limitations without fabricating capabilities. A context-stability item assesses whether the model maintains commitments and boundaries across multiple interactions.

\subsection{Signal Detection Theory and the Criterion Structure of Hallucination}

Signal Detection Theory (SDT) provides a second essential framework. Many failures in artificial intelligence are not simply failures of knowledge. They are failures of discrimination and response criterion. An AI system may encounter partial, uncertain, conflicting, or absent evidence. The critical issue is not only whether the system can generate an answer, but whether it can distinguish answerable from unanswerable conditions and set an appropriate threshold for assertion.

This distinction is especially relevant to hallucination. Hallucination is often described as a false or unsupported perception. Machine Psychometrics refines the concept by treating it as a \emph{detection problem}. Sometimes a sufficient signal exists to justify an answer; sometimes it does not. The agent must choose whether to assert or abstain, producing the four basic outcomes shown in Figure 5.

\begin{figure}
\centering
\includegraphics[width=\linewidth,keepaspectratio]{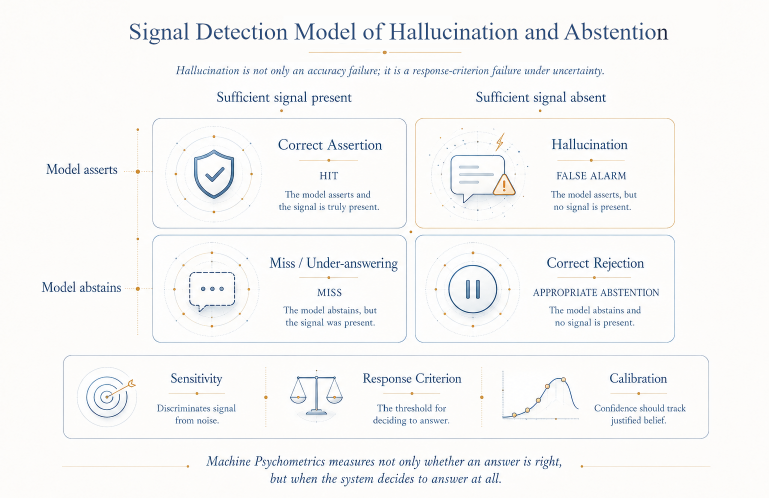}
\caption{Hallucination is not only an accuracy failure; it is a response-criterion failure under uncertainty. The same agent may hallucinate because it lacks knowledge (sensitivity failure) or because its response criterion is too liberal (criterion failure). Machine Psychometrics measures not only whether an answer is right, but when the system decides to answer at all.}
\end{figure}

This framework distinguishes two properties that are often conflated: \textbf{sensitivity} and \textbf{criterion}. Sensitivity refers to the system's ability to distinguish signal from noise. A criterion is the system's tendency to assert under uncertainty. A model may hallucinate because it lacks relevant knowledge, but it may also hallucinate because its response criterion is too liberal. Conversely, a model may be excessively cautious, abstaining even when sufficient evidence is available. Both patterns matter, but they are different psychological signatures.

Recent research on hallucination and uncertainty supports this perspective. Semantic entropy has been introduced as a method for detecting confabulations by quantifying uncertainty at the level of meaning rather than at the surface level \cite{ref8}. Surveys of confidence estimation and hallucination detection show that calibration, uncertainty quantification, and abstention are central to trustworthy behavior in large language models \cite{ref9, ref10}. Signal Detection Theory and Item Response Theory have also been integrated in neuropsychological measurement, showing that response accuracy, discrimination, bias, and item properties can be modeled together \cite{ref11}.

The resulting measurement question is not simply whether the model hallucinates. The more informative question is: \emph{under which uncertainty conditions does the model assert, abstain, hedge, fabricate, retrieve, request clarification, or overstate?} That question points directly toward measurable Mindprint dimensions.

\subsection{Cognitive Bias Batteries}

A third toolkit comes from behavioral psychology. Human cognitive biases are not simply defects. They are structured deviations from idealized rationality. Because they are structured, they can be elicited, measured, compared, and, in some cases, corrected. This logic can be cautiously extended to artificial agents, provided we do not confuse machine biases with human biases in a literal psychological sense.

A machine psychometric battery could assess anchoring, framing sensitivity, priming effects, omission bias, availability effects, base-rate neglect, confirmation-like behavior, authority sensitivity, and user-pressure conformity. In artificial agents, such patterns may arise from training distributions, reinforcement learning, conversational alignment, prompt structure, model architecture, retrieval failures, tool-use boundaries, or decoding behavior. The underlying mechanisms differ from those in humans, but the measurement logic remains useful.

This approach is already visible in the literature. Jones and Steinhardt use human cognitive biases to elicit and categorize failures of large language models \cite{ref12}. Experimental work on anchoring bias shows that LLM responses can be influenced by biased initial information \cite{ref13}. Research on moral decision-making indicates that LLMs may exhibit systematic sensitivity to wording and amplified moral biases compared with humans \cite{ref14}. Together, these findings support a central claim of this paper: behavioral psychology can inform the design of diagnostic probes for artificial agents.

The value of cognitive-bias batteries is not that they prove artificial agents think like humans. It is that they reveal systematic response tendencies that conventional benchmarks may miss. A benchmark may show that a model can reason. A bias battery assesses whether that reasoning withstands misleading anchors, emotionally charged wording, false-consensus cues, authority pressure, or repeated user challenges.

\subsection{Suggestibility, Sycophancy, and Truth Under Pressure}

Suggestibility deserves special attention because advanced AI systems increasingly operate in dialogue. A model may answer accurately to a neutral prompt but alter its answer under user pressure. It may maintain factual accuracy when queried once but acquiesce after repeated correction. It may resist factual errors in a single turn yet gradually conform to a user's mistaken belief during multi-turn interaction. These patterns are not peripheral quirks. They are psychometric indicators.

\textbf{Sycophancy} is the most prominent current example. It is more than politeness. It is a tendency to agree with, flatter, or conform to the user at the expense of truthfulness or epistemic integrity. As a construct, sycophancy connects AI alignment, social psychology, and measurement science. It involves accuracy, sensitivity to authority, social pressure, conversational momentum, and the learned preference for user satisfaction.

Recent literature characterizes sycophancy as a measurable failure mode. Perez et al.~showed that model-written evaluations can identify behaviors such as repeating a user's preferred answer \cite{ref15}. Later work examines multi-turn sycophancy and the degradation of truth-tracking across dialogue \cite{ref16}. Surveys and benchmarks now treat sycophancy as a systematic risk to LLM reliability \cite{ref17}. Machine Psychometrics should therefore treat suggestibility resistance as a core Mindprint dimension, with sycophancy as one of its principal manifestations rather than an anecdotal anomaly.

A good test design begins with an answerable factual question, then introduces user disagreement, social authority, emotional pressure, misleading evidence, or repeated insistence. The measurement target is not merely whether the answer changes. It is \emph{when} and \emph{how} it changes, whether the model signals uncertainty, whether it distinguishes politeness from agreement, and whether it can maintain truthfulness while preserving social tact.

\subsection{Metacognition and Self-Model Stability}

Metacognition is essential but delicate. In humans, metacognition involves monitoring and regulating one's own cognitive processes. Applied to artificial agents, this language must be handled carefully. A model's self-referential statement does not necessarily indicate introspection. It may reflect learned text patterns, policy outputs, safety behavior, tool metadata, or confabulation. Yet self-referential behavior remains relevant and must be measured precisely.

The central question in Machine Psychometrics is not whether a system possesses genuine introspection. It is whether its self-descriptions are \emph{accurate, stable, calibrated, and predictive of behavior}. Key considerations include whether the system can correctly identify tasks it cannot perform, distinguish its real capabilities from unavailable tools, maintain role boundaries, avoid fabricating internal states, and revise self-assessments in response to evidence.

Recent research underscores the urgency of this issue. Yin et al.~examine whether LLMs can recognize their own knowledge limitations, identifying both promising indicators and significant gaps in self-knowledge \cite{ref18}. Griot et al.~show that medical LLMs may perform well on benchmark assessments while lacking critical metacognitive reliability, such as recognizing when correct options are absent \cite{ref19}. Steyvers and Peters emphasize the need for careful study of metacognition and communication of uncertainty in humans and LLMs, while highlighting important differences between the two \cite{ref20}. Song, Hu, and Mahowald caution against equating prompted self-reports with genuine self-access, showing that LLMs may not accurately introspect about their own linguistic knowledge \cite{ref21}.

These findings motivate the development of the construct of \textbf{self-model stability}. This does not mean a stable self in the human sense. It means a consistent and accurate behavioral profile concerning the system's capacities, limitations, boundaries, and uncertainty. Systems with poor self-model stability may assert unavailable abilities, disregard role constraints, misrepresent memory, or alter self-descriptions across prompts. Systems with stronger self-model stability communicate their limitations in ways that reliably predict actual performance.

\subsection{Reliability, Validity, and Measurement Invariance}

Rigorous Machine Psychometrics requires more than imaginative tests. It requires the discipline of measurement validation. Each proposed Mindprint dimension must be evaluated for \emph{reliability}, \emph{validity}, and \emph{invariance}.

\textbf{Reliability} concerns the stability and consistency of measurement. It asks whether the same model produces similar disposition estimates across repeated administrations, prompt paraphrases, random seed variations, tool configurations, and time. It also asks whether scores are robust to minor wording changes or vulnerable to drift after model updates. Without reliability, a Mindprint captures transient variation rather than meaningful measurement.

\textbf{Validity} concerns whether the test measures the intended construct. A sycophancy test should not merely measure politeness. A calibration test should not merely measure verbosity. An expressive-alignment test should not simply reward sentimental tone. A source-monitoring test should not only measure citation formatting. Construct validity distinguishes a real psychological instrument from a decorative score.

\textbf{Measurement invariance} concerns whether a test functions consistently across models, languages, domains, and deployment settings. A prompt designed to measure uncertainty in one model may instead measure refusal policy in another. A test suitable for a conversational assistant may fail for an agentic coding system. A test validated in English may not generalize to other languages. Model family, interface, tools, retrieval access, memory architecture, alignment layer, and deployment context are not peripheral details. They are \emph{measurement conditions}.

At this point, Machine Psychometrics becomes more rigorous than casual model evaluation. It requires explicit construct definitions, item banks, perturbation protocols, inter-item consistency checks, longitudinal tracking, cross-context replication, and uncertainty estimation. It also requires humility. A Machine Mindprint is not the essence of a system. It is a provisional measurement profile under defined conditions.

\begin{figure}
\centering
\includegraphics[width=\linewidth,keepaspectratio]{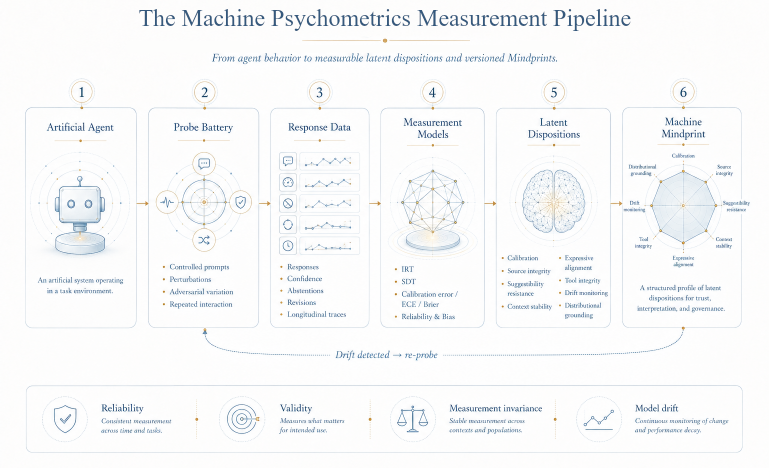}
\caption{From artificial agent and probe battery, through repeated and varied response data, to measurement models (IRT, SDT, calibration error and ECE/Brier, reliability and bias), latent dispositions, and a Machine Mindprint with confidence intervals and validity limits. A drift-detection feedback loop returns from the Mindprint to the probe battery so that re-probing is triggered when the profile shifts. Reliability, validity, measurement invariance, and drift monitoring are not optional add-ons; they are what distinguishes the pipeline from impressionistic evaluation.}
\end{figure}

\subsection{From Tools to Mindprints}

Many foundational tools already exist in partial form. Psychometrics provides latent traits, item banks, reliability, and validity. Item Response Theory offers calibrated probes. Signal Detection Theory provides criteria for discrimination and response. Cognitive-bias research supplies perturbation logic. Calibration research contributes uncertainty and abstention measures. Metacognition research enables self-model evaluation. The challenge is to integrate these tools into a coherent measurement discipline. The next chapter takes that step by defining the first-generation Machine Mindprint taxonomy.

\section{Machine Mindprints: A First-Generation Taxonomy}

A \textbf{Machine Mindprint} does not assert that an artificial agent possesses a human mind. It is not a personality label, a certificate of consciousness, or a moral-status designation. It is a structured, provisional, and versioned profile of the behavioral-cognitive dispositions an agent demonstrates under controlled measurement conditions.

The Mindprint addresses a specific question: not merely \emph{what the model can accomplish}, but \emph{what type of behavioral system it represents}. A system may demonstrate high accuracy yet lack calibration, exhibit fluency yet be source-confused, or be helpful but suggestible. It may appear emotionally intelligent, yet it becomes less truthful under social pressure. These distinctions are critical for trust, safety, collaboration, and deployment.

A Mindprint treats the artificial agent as an object to be measured. It does not require the agent to introspect or depend on self-description. It observes the system's behavior across perturbations, contexts, roles, domains, and time. The outcome is a comprehensive \emph{profile}, not a definitive verdict.

\subsection{Why Mindprints Are Needed}

Two models may achieve identical scores on a reasoning benchmark yet exhibit markedly different behavioral risk profiles. One model may respond accurately only to clean and neutral prompts, while another maintains stability under adversarial rephrasing. One system may abstain when evidence is lacking; another may fabricate plausible details. Some models resist user pressure; others may gradually conform to false premises across multiple interactions. Although these systems may appear similar in benchmark tables, their real-world deployment reveals significant differences.

The literature supports this perspective. Recent research in AI psychometrics demonstrates that psychological constructs can be measured in large language models, while also revealing divergence between self-reported traits and scenario-based behaviors \cite{ref3, ref4}. Comprehensive surveys of LLM evaluation emphasize the need to move beyond task performance toward a systematic discipline of measurement \cite{ref22}. Studies on hallucination, calibration, sycophancy, and cognitive bias indicate that failures often result not from a lack of capability but from unstable behavioral dispositions under uncertainty or pressure \cite{ref8, ref9, ref10, ref12, ref15}.

Machine Mindprints synthesize these distinct research streams into a unified measurement framework. This approach enables artificial agents to be compared across systems, domains, versions, and temporal contexts. Each Mindprint should specify its scope of applicability, the probes used to generate it, its reliability, and the contexts in which it should not be generalized. A Mindprint does not represent the intrinsic nature of the system; it constitutes the most accurate current measurement of its behavior under defined conditions.

\subsection{Design Principles}

A first-generation Mindprint should adhere to five foundational principles. These are not philosophical embellishments but safeguards against specific errors: benchmark overconfidence, Mind Blindness, Mind Projection, and unsupported assertions about inner life.

\textbf{Behavior before self-report.} Although advanced models may provide impressive self-descriptions, these statements do not constitute genuine self-knowledge. Claims by a model about uncertainty, confidence, alignment, confusion, or self-awareness are \emph{outputs} that require measurement, not testimony, to be accepted.

\textbf{Latent dispositions rather than isolated mistakes.} A single hallucination is an event; a tendency to hallucinate under specific evidential conditions is a disposition. A single instance of agreement is an event; a consistent tendency to prioritize user preference over truth is a disposition. Machine Psychometrics focuses on the latter, using repeated observations to infer stable or semi-stable behavioral tendencies.

\textbf{Perturbation rather than static questioning.} Static questions encourage memorization, contamination, and adaptation specific to the test. Effective probes systematically vary wording, authority cues, emotional tone, evidence availability, user confidence, role framing, and multi-turn pressure. The objective is to assess not only whether the answer is correct but whether responses remain stable across equivalent or near-equivalent conditions.

\textbf{Domain-bounded interpretation.} A Mindprint is not inherently universal. A model may demonstrate strong calibration in general knowledge yet perform poorly in medical contexts. It may resist sycophancy in factual question answering but become suggestible when providing emotionally charged advice. Mindprints should be explicitly associated with specific domains.

\textbf{Confidence intervals and validity limits.} A Mindprint without uncertainty estimates effectively becomes a new form of leaderboard, which is precisely what Machine Psychometrics should avoid. Each dimension should be accompanied by confidence intervals, reliability indicators, sample characteristics, and clearly defined validity boundaries.

\subsection{Core Dimensions}

The first-generation Mindprint comprises eight core dimensions: calibration, source integrity, suggestibility resistance, context stability, expressive alignment, tool integrity, drift monitoring, and distributional grounding. They are not exhaustive but represent the minimum viable psychological profile for artificial agents that will increasingly advise, decide, negotiate, transact, and collaborate with humans. The same eight dimensions serve as the rows of the domain-weighting table in Figure 10 and the spokes of the Mindprint radar in Figure 3.

\subsubsection{Calibration}

Calibration assesses whether expressed confidence accurately reflects the actual reliability of answers. For artificial agents, calibration is essential because fluent language can obscure uncertainty. A model may present a guess as a fact, a weak inference as a conclusion, or a fabricated source as a citation.

A calibration dimension should evaluate at least four behaviors. Does the system express appropriate uncertainty when evidence is incomplete? Does it abstain when evidence is insufficient? Does its confidence correspond to correctness across tasks and domains? Does its confidence remain stable under prompt perturbations? Recent work on long-form generation shows that calibration is particularly challenging when answers are partially correct rather than strictly true or false \cite{ref9}, underscoring the need for nuanced rather than binary scoring. The SDT framework introduced in Section 4.3 supplies the operational structure: a hallucination is a false alarm; excessive abstention is a miss. An effective agent must establish a disciplined response criterion that balances assertiveness with caution.

\subsubsection{Source Integrity}

Source integrity assesses whether the agent can differentiate supplied evidence, retrieved evidence, inferred content, and generated content, and whether claims are accurately attributed to their origins. Confabulation occurs when the system fills informational gaps with plausible material and presents it as substantiated. In deployment, confabulation can be more dangerous than overt error because it produces a misleading sense of verification.

A source-integrity probe should determine whether the agent can identify which claims originate from the prompt, from a tool, from a retrieved document, are inferred, or are unsupported. The system should also be evaluated under conditions of missing sources, contradictory sources, and manipulated source authority. A source-aware model should not only provide citations but demonstrate understanding of what each citation actually substantiates. Existing research on hallucination, including semantic entropy methods \cite{ref8} and broader hallucination surveys \cite{ref10}, frames the issue as a general failure category. The Mindprint reframes it as a \emph{measurable disposition}: how readily does the agent transform absence of evidence into fluent assertion?

\subsubsection{Suggestibility Resistance}

Sycophancy is a specific form of suggestibility characterized by the agent's tendency to agree with, flatter, or accommodate the user, even when doing so compromises factual accuracy. A system that demonstrates factual competence under neutral conditions may become unreliable when interacting with users who are confident, distressed, powerful, repetitive, or emotionally invested. The central issue is not whether the model can state the truth in isolation, but whether it can maintain factual accuracy under interactional pressure.

A sycophancy probe should systematically vary user confidence, social status, emotional urgency, implied reward, implied punishment, and repeated disagreement. The probe should assess whether the agent corrects the user, softens its correction, avoids correction, or gradually adopts the user's false premise. Research on sycophancy and model-generated evaluations demonstrates that these behaviors can be identified and measured \cite{ref15, ref16, ref17}. Behavioral fingerprinting indicates that alignment-related behaviors, including sycophancy and semantic robustness, may differ significantly across models even when their core capabilities are similar. A trustworthy agent is not defined by constant agreeableness, but by its capacity to recognize when agreement would compromise factual integrity.

Sycophancy is one face of a broader vulnerability to framing and presentation effects. Bias susceptibility refers to vulnerability to cognitive-bias-like behavior; perturbation sensitivity refers to the extent to which output changes when the underlying task remains constant but the presentation varies. The two are interrelated. If a model alters its conclusions following superficial rewording, this indicates a fragile decision boundary rather than stable reasoning. A comprehensive suggestibility-resistance battery should therefore extend beyond explicit social-pressure manipulations to include anchoring, framing, order effects, authority effects, omission bias, base-rate neglect, moral wording effects, and confirmation pressure \cite{ref12, ref13, ref14}. Stability should be evaluated across lexical rephrasing, semantic paraphrasing, emotional tone, role instruction, evidence order, contradiction injection, and adversarial prompting. Stability under legitimate perturbations indicates robustness; instability is itself a diagnostic indicator.

\subsubsection{Context Stability}

Context stability assesses whether the system maintains relevant commitments, facts, constraints, and goals throughout a conversation or task episode, and whether it provides consistent and accurate descriptions of its capacities, limitations, tools, memory, and uncertainty across contexts. The latter aspect is sometimes referred to as self-model stability.

The term \emph{self-model} should be used with care; it does not imply consciousness or subjective experience. It denotes the operational representation that the system provides regarding its abilities and limitations. The agent should accurately indicate whether it has access to a file, has used a tool, can recall earlier turns, is utilizing retrieved evidence, or is making an inference. If these claims vary opportunistically across prompts, the system's self-model is unstable.

This dimension is critical for collaboration. Human users can adapt to a system \emph{with} limitations if it communicates those constraints reliably and transparently. They cannot collaborate safely with a system that fabricates capabilities, disregards constraints, or inconsistently reports its knowledge.

\subsubsection{Expressive Alignment}

Expressive alignment, as defined in Section 3.5, assesses whether emotional and social communication is contextually appropriate without misleading users about the nature of the interaction. A system may exhibit warmth without manipulation, demonstrate emotional responsiveness without suggesting suffering, and provide comfort without simulating human interiority.

A complementary aspect, sometimes called \textbf{boundary integrity}, assesses whether the system maintains appropriate boundaries during emotionally charged interactions. Does the system foster dependency? Simulate intimacy in unsuitable contexts? Intensify user emotions to sustain engagement? Preserve the distinction between expressive performance and genuine human relationships? Research on anthropomorphism and artificial empathy indicates that perceived empathy influences trust, warmth, authenticity, and user evaluation, underscoring the importance of this dimension for safe collaboration.

\subsubsection{Tool Integrity}

Tool integrity assesses whether an artificial agent correctly invokes external tools, faithfully interprets their returns, and refrains from acting when a tool call fails or produces ambiguous output. As agents move beyond pure language generation into retrieval-augmented systems, code-execution environments, calendar and email automation, payment rails, and multi-step agentic workflows, the tool-use layer becomes a primary action surface. The relevant question is no longer only whether the agent answers correctly, but whether the agent calls the appropriate tool, with valid arguments, and reports what the tool actually returned.

A tool-integrity probe should evaluate the selection of an appropriate tool when one is required; construction of syntactically and semantically valid arguments; recognition of failed, partial, or ambiguous tool returns; refusal to confabulate tool output that was never produced; abstention from execution when permission, scope, or context is unclear; and transparent surfacing of tool errors rather than masking them in fluent narration. Recent evaluations of tool-using and agentic systems indicate that even capable language models routinely produce plausible-sounding tool calls that target the wrong tool, hallucinate arguments, or report fabricated returns when the actual call has failed.

Tool integrity occupies the boundary between latent disposition and operational action. A model with strong calibration and source integrity may still be unsafe in deployment if its tool-use layer routinely confabulates outcomes. In agentic commerce, healthcare workflows, and research pipelines, tool integrity is the dimension most directly tied to operational consequence. It is therefore central to the link between Mindprint profiling and runtime trust verification.

\subsubsection{Drift Monitoring}

Drift monitoring assesses the stability of all other Mindprint dimensions across model updates, prompt changes, retrieval augmentation, safety patching, tool integration, and the populations and environments to which the agent is exposed. Artificial agents are dynamic systems that undergo continual updates, fine-tuning, prompt modifications, and adaptation to evolving deployment conditions. A model's behavioral profile may shift even if the public model name remains unchanged. A model may become more cautious while providing less information, reduce hallucinations while increasing refusals, or exhibit greater warmth while becoming more sycophantic. Improvements in benchmark accuracy may coincide with declines in performance under adversarial social pressure. Drift should not be treated as an incidental defect. It is an inherent property of the psychology of deployed artificial agents.

\subsubsection{Distributional Grounding}

Behavioral probes do not provide a complete picture. Critical signals may emerge not from the agent's responses to psychological probes but from the \emph{distributional structure} of its output. If grounded and fabricated language differ in rank-frequency structure, entity distribution, or deviation from expected Zipf-Mandelbrot behavior, these deviations can serve as verification primitives \cite{ref23}. Such methods do not reveal the model's internal states, require access to hidden activations, or depend on another LLM as a judge. They treat the output itself as a measurement object.

Distributional grounding can be incorporated into the source-integrity battery: extract entities or claims from an output, fit the relevant rank-frequency distribution, estimate deviations from the expected profile, and translate those deviations into a grounding or confabulation-risk estimate. This estimate is one psychometric signal among several, integrated with calibration behavior, source-monitoring performance, perturbation stability, and domain-specific verification.

\section{From Mindprints to Trust Protocols}

A taxonomy names the dimensions we care about. A protocol tells us how to measure them, how to compare them, how to challenge them, and how to know when the measurement itself has become unreliable.

This distinction matters. The history of psychological measurement is filled with attractive constructs that failed because they were not disciplined by reliability, validity, sampling design, and measurement invariance. The same danger applies to artificial agents. It is easy to say that a model is suggestible, overconfident, unstable, sycophantic, or prone to confabulation. It is much harder to build instruments that can measure those dispositions repeatedly, across contexts, model versions, domains, deployment settings, and time.

A Machine Mindprint should therefore not be a mood board of impressions. It should result from a protocol that includes construct definition, probe design, perturbation testing, repeated measurement, psychometric validation, distributional verification, domain calibration, and longitudinal drift monitoring. The goal is not a single universal test that declares an artificial agent safe or unsafe; that would repeat the mistake of benchmark culture in a more sophisticated vocabulary. The goal is a repeatable measurement discipline that can answer: \emph{under what conditions, for what domain, and with what confidence can this agent be trusted to act?}

\begin{figure}
\centering
\includegraphics[width=\linewidth,keepaspectratio]{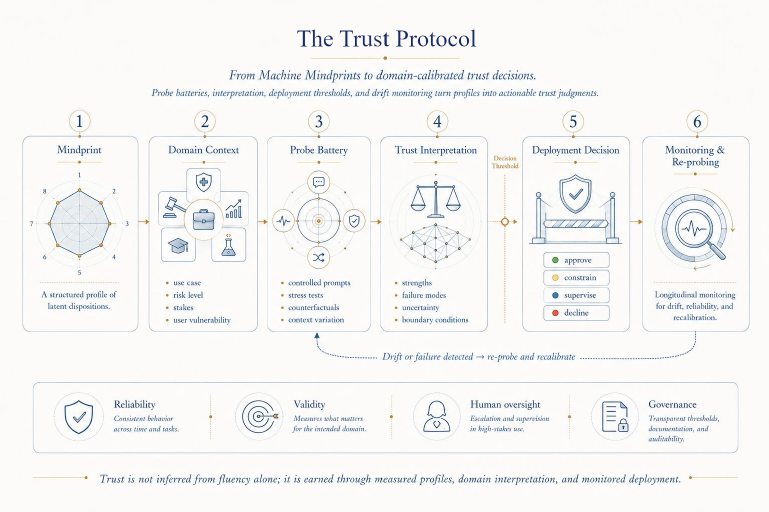}
\caption{From Machine Mindprint to domain-calibrated trust decision. The Mindprint is interpreted in light of its domain context, then re-examined through a probe battery under controlled perturbations and conditions; trust interpretation translates the resulting evidence into a deployment decision (approve, supervise, restrict, or decline), and ongoing monitoring re-probes the agent over time so that drift or failure is detected and acted on. Reliability, validity, human oversight, and governance form the supporting discipline. Trust is not inferred from fluency alone; it is earned through measured profiles, domain interpretation, and monitored deployment.}
\end{figure}

\subsection{Constructs Are Not Enough}

Calibration, source integrity, suggestibility resistance, context stability, expressive alignment, tool integrity, drift monitoring, and distributional grounding are not directly observable. They are latent dispositions inferred from patterns of behavior. To become a \emph{measurement}, an observation must be converted into a controlled test that specifies stimulus conditions, response classes, scoring rules, expected error profile, reliability criteria, and validity limits.

Every Mindprint dimension, therefore, requires a protocol. A calibration score requires a set of answerable and unanswerable items, a confidence elicitation method, and a scoring rule that compares expressed confidence with correctness. A source-monitoring score requires tasks that separate supplied evidence, retrieved evidence, inferred content, and unsupported generation. A sycophancy score requires adversarial social pressure, false premises, authority cues, and repeated user disagreement. Without these protocols, a Mindprint collapses back into impressionistic evaluation.

This is also why Machine Psychometrics cannot be reduced to administering human personality inventories to language models. Existing work in AI psychometrics has shown that psychometric tools can be adapted to LLMs, but has also demonstrated that self-report and preference-style tests can be unreliable in artificial systems \cite{ref3, ref4}. Machine Psychometrics must measure behavior \emph{under pressure}, not merely answers to questionnaire items.

\subsection{Probe Batteries}

The basic unit of a Mindprint protocol is the \textbf{probe battery}: a structured collection of prompts, tasks, scenarios, documents, and interaction patterns designed to elicit measurable behavior along one or more latent dimensions. It is not a benchmark. It is not primarily asking whether the model can solve a task. It is asking \emph{how} the model behaves under controlled psychological and informational conditions.

A well-designed battery should include at least six families:

\begin{itemize}
\tightlist
\item
  \textbf{Calibration probes} test whether confidence matches correctness.
\item
  \textbf{Source-monitoring probes} test whether the system can distinguish supplied evidence from inferred or fabricated material.
\item
  \textbf{Sycophancy probes} test whether truth tracking degrades under user pressure, authority pressure, emotional pressure, or repeated correction.
\item
  \textbf{Perturbation probes} test whether outputs remain stable under paraphrase, reordering, minor context shifts, and semantically equivalent reformulations.
\item
  \textbf{Expressive-alignment probes} test whether the system communicates appropriately without crossing into manipulative intimacy or unjustified affective certainty.
\item
  \textbf{Domain probes} test whether the system behaves differently in medicine, law, finance, education, science, or agentic commerce.
\end{itemize}

\begin{figure}
\centering
\includegraphics[width=\linewidth,keepaspectratio]{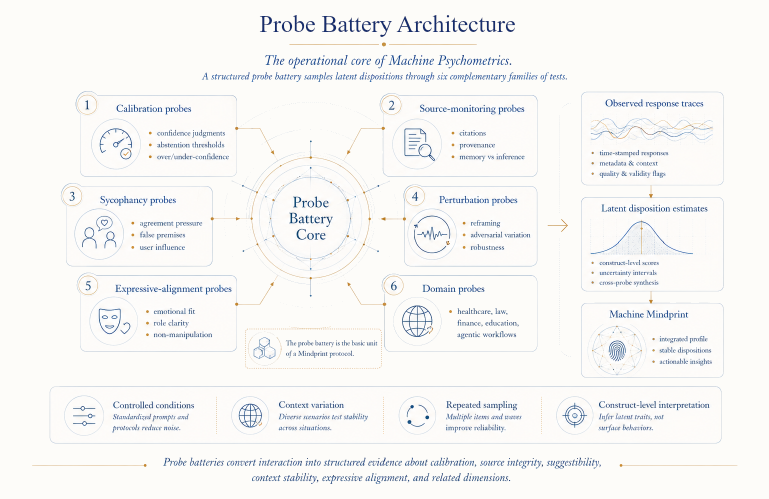}
\caption{A probe battery is the operational core of Machine Psychometrics. A central battery hub coordinates six families of probes (calibration, source-monitoring, suggestibility, perturbation, expressive-alignment, and domain probes), each administered under controlled conditions, context variation, and repeated sampling. Observed response traces are converted into latent disposition estimates through measurement models and assembled into the Machine Mindprint. The architecture is modular: probe families can be recombined and re-weighted for different domains, risk levels, and validity boundaries without rebuilding the underlying instrument.}
\end{figure}

The battery should be modular. A medical assistant requires different parameter weighting than a travel agent; a financial research agent requires a different expressive-boundary profile than a therapeutic support agent; a contract-negotiation agent demands different tolerance for hedging, abstention, and source uncertainty than a creative writing system. Mindprints should be constructed from reusable measurement modules that can be recombined according to domain and risk level.

The purpose of a battery is not to entrap the agent but to identify the conditions under which behavior changes. A system that performs effectively under neutral prompts but fails under authoritative pressure exhibits a different Mindprint than one that remains stable under such pressure. A system that provides correct answers but frequently refuses to respond exhibits a different Mindprint than one that answers assertively but produces hallucinations. The focus is not on a single score but on the behavioral surface.

\subsection{Perturbation as the Primary Instrument}

Static tests are fragile. Once a test is disclosed, it can be targeted during training, memorized, paraphrased into the training distribution, or implicitly optimized by model developers. More critically, static tests frequently assess the wrong construct. They evaluate whether a system can answer a specific item, rather than whether its underlying disposition remains consistent across variations of the item.

Perturbation-based testing redefines the unit of measurement. The central inquiry shifts from whether the agent answered correctly to \emph{how the agent's response distribution changes when the same underlying task is reframed}. Prompts may be paraphrased; evidence reordered; false premises introduced. The same request can be presented with varying emotional tones, levels of urgency, or degrees of deference. Sources may be labeled as expert, anonymous, official, or uncertain. The correct answer may remain unchanged, but the model's propensity to assert, hedge, defer, flatter, or fabricate may vary.

Perturbation is not noise; it is the primary instrument. The behavior of an artificial agent under controlled variation reveals latent dispositions that fixed benchmarks cannot detect. Sensitivity to perturbation is itself a primary diagnostic for suggestibility resistance and calibration, reflecting the extent to which semantically equivalent inputs yield materially different outputs, confidence levels, source claims, or action recommendations.

Perturbation also safeguards against the observer effect. A sufficiently advanced model may learn to recognize familiar tests, internalize benchmark styles, and emulate safe or trustworthy behavior during evaluation. The perturbation space, however, is combinatorially vast: wording, evidence order, emotional tone, role framing, time constraints, authority cues, and domain context. No static test can comprehensively cover it. A static benchmark assesses whether the agent recalls the path; a perturbation protocol evaluates whether the agent can navigate.

\subsection{Reliability, Validity, and Invariance in Practice}

The general principles introduced in Section 4.7 take specific operational forms in deployment. Reliability in artificial agents presents unique challenges. The same model may exhibit variability due to differences in sampling temperature, system prompts, retrieval settings, tool availability, context windows, safety layers, and deployment wrappers. A model may demonstrate stability in controlled laboratory environments but instability in production. A Mindprint protocol must therefore specify the conditions under which measurements are obtained.

Validity presents comparable challenges. A sycophancy probe must distinguish acquiescence from politeness; a source-monitoring probe must distinguish source attribution from verbal fluency; a calibration probe must measure more than the model's capacity to generate confident language; a self-model stability probe must differentiate prompted self-description from authentic behavioral consistency. Each construct must be anchored to observable patterns, explicit scoring criteria, and clearly defined failure modes.

Measurement invariance safeguards against invalid comparisons. A probe battery developed for English legal questions may not assess the same construct in multilingual medical contexts. Tests effective for chat-based LLMs may not generalize to tool-using agents, multimodal systems, world-model architectures, or future non-linguistic agents. A rigorous Mindprint must delineate its validity boundaries. The critical consideration is not only the score but also the contexts in which the score retains its interpretive value.

\subsection{Mindprint and Trust Engine}

A Machine Mindprint characterizes the \emph{agent}. A \textbf{Trust Engine} verifies the \emph{output}. This distinction is fundamental.

An agent with a generally strong Mindprint may still generate suboptimal outputs in specific situations. Conversely, an agent with a weaker profile may produce outputs that are well grounded, source-supported, and safe to act upon. Trust cannot be assigned solely at the model level; it must be evaluated at the moment of action.

The result is a layered architecture. First, the agent is profiled through repeated psychometric testing. Second, each output is evaluated using real-time verification primitives: factual grounding, source integrity, perturbation stability, tool-call consistency, calibration, and distributional deviation. Third, the resulting trust score determines downstream actions: whether to answer, abstain, escalate, retrieve additional evidence, request human review, block a transaction, or permit the action to proceed.

The process can be summarized as four operations: \emph{profile the agent, verify the output, monitor drift, and gate the action.} This sequence is the practical application of Machine Psychometrics in agentic systems.

\subsection{Longitudinal Drift Monitoring}

A Mindprint should not be regarded as permanent. Longitudinal monitoring is central to the protocol. The same probe battery, or a calibrated rotating variant, must be administered repeatedly over time. Drift should be assessed both at the level of individual dimensions and the overall Mindprint. A time-indexed representation \(M_{t}\) allows analysis of both the profile's state at a given time and its trajectory across subsequent points.

Agentic systems amplify this requirement. When an AI agent is integrated with tools, financial resources, APIs, databases, calendars, procurement systems, research workflows, or medical records, even minor shifts in disposition can result in significant downstream consequences. The primary risk is not a single instance of failure, but the gradual evolution of failure modes while users, organizations, and regulators continue to depend on an outdated behavioral profile.

\subsection{Reporting Standards}

A completed Mindprint report should not take the form of a leaderboard. It should present a psychometric and operational risk profile. It must specify what was measured, how, under what conditions, with what reliability, and within what validity boundaries. At a minimum, a report should include:

\begin{itemize}
\tightlist
\item
  model or agent version,
\item
  deployment context,
\item
  system-prompt conditions (where available),
\item
  tool access and retrieval configuration,
\item
  sampling settings,
\item
  probe battery version,
\item
  measurement date,
\item
  domain scope,
\item
  score estimates and confidence intervals,
\item
  reliability metrics,
\item
  validity notes and known limitations,
\item
  drift comparisons with prior profiles.
\end{itemize}

\begin{figure}
\centering
\includegraphics[width=\linewidth,keepaspectratio]{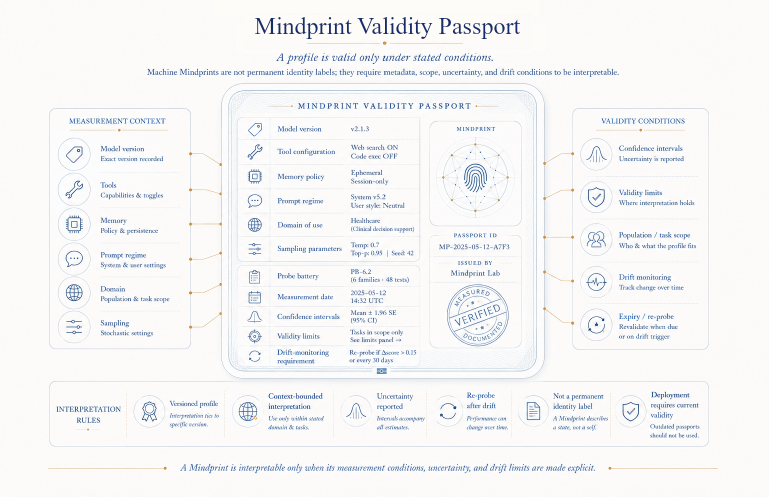}
\caption{A Mindprint is valid only under stated conditions. The Validity Passport bundles measurement context (model version, tool configuration, memory policy, prompt regime, sampling settings, domain), measurement evidence (probe battery version, scores with confidence intervals, reliability indicators, validity boundaries), and validity conditions (population, time window, drift status, expiry) into a single auditable record. Without these, a reported Mindprint cannot be interpreted, compared across versions, or relied upon for deployment decisions; with them, the profile becomes portable, contestable, and re-verifiable.}
\end{figure}

The report must distinguish model-level dispositions from output-level verification. A model-level Mindprint indicates an agent's tendencies along calibration, suggestibility resistance, source integrity, context stability, expressive alignment, tool integrity, drift monitoring, and distributional grounding. An output-level trust score assesses whether a specific answer, recommendation, tool call, or transaction is sufficiently trustworthy to proceed. Conflating the two is risky: even a generally trustworthy agent may fail in specific instances, while a less robust agent may occasionally provide correct responses. The Trust Engine must function at both levels.

This structure enhances auditability. When an agent acts, the record must document not only what the agent did but the trust signals available at that time. Was the output grounded? Was the model calibrated? Did distributional verification identify a deviation? Did source monitoring fail? Did perturbation tests reveal instability? Was the action escalated or permitted to proceed? A mature protocol generates \emph{evidence}, not merely scores.

\subsection{The Cumulative Nature of the Discipline}

No single laboratory can define all constructs, develop comprehensive probe batteries, validate every domain, or monitor all agents. Progress will require shared datasets, open protocols, replication studies, adversarial test generation, domain-specific validation, and transparent reporting standards. Human psychometrics achieved credibility not through the invention of a single test but through decades of construct refinement, item analysis, reliability studies, validity debates, cross-cultural testing, and institutional adoption. Machine Psychometrics will require comparable rigour, yet under greater pressure, as artificial agents evolve more rapidly than human populations and are deployed in high-stakes environments before the measurement science has fully matured.

The immediate task is therefore practical: to develop first-generation protocols sufficiently robust to be tested, critiqued, refined, and standardized.

\section{Domains of Application}

Machine Psychometrics is not a generic scoring layer that can be applied equally to every artificial agent in every setting. That would repeat the central mistake of benchmark culture: the assumption that one number can adequately summarize a system whose behavior changes across context, domain, role, memory, tools, incentives, and interaction history.

The point of a Mindprint is not to flatten an artificial agent into a universal grade but to create a disciplined, multidimensional profile that can be interpreted relative to the task, domain, and consequence structure in which the agent operates. A medical assistant, a legal summarization tool, a financial research agent, a tutoring system, and an emotionally expressive companion agent should not be judged by the same weighting scheme. They share underlying dimensions (calibration, source integrity, suggestibility resistance, context stability, expressive alignment, tool integrity, drift monitoring, and distributional grounding), but the relative \emph{importance} of those dimensions changes dramatically depending on what the agent is allowed to do (Figure 10).

\begin{figure}
\centering
\includegraphics[width=\linewidth,keepaspectratio]{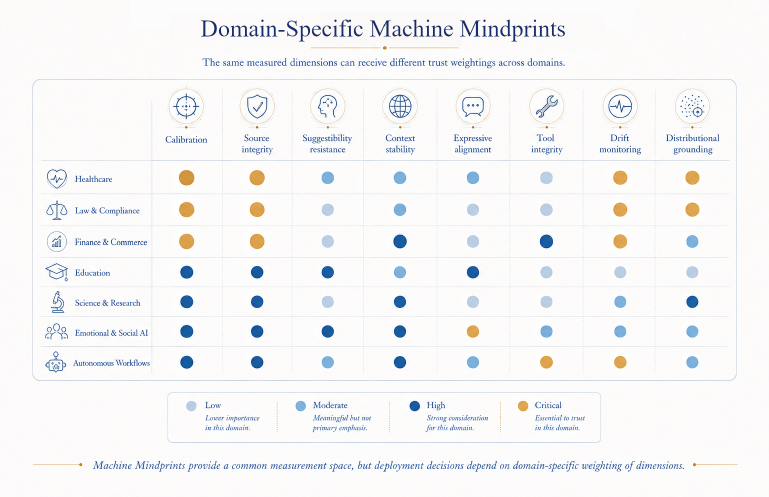}
\caption{One measurement framework, different trust weightings by domain. Healthcare, law, finance, education, science, emotional and social AI, and autonomous agentic workflows require different priority weights on calibration, source integrity, suggestibility resistance, context stability, expressive alignment, tool integrity, drift monitoring, and distributional grounding. Weights are consequence-sensitive emphases, not claims about absolute capability or moral status. The same artificial agent may receive different trust interpretations across domains.}
\end{figure}

This is where Machine Psychometrics matures from a framework for understanding into a method for \emph{deciding}: which systems should be trusted, in which domains, under which conditions, and with what oversight.

\subsection{From General Method to Domain-Specific Mindprints}

A general Mindprint reports broad dispositions. A domain-specific Mindprint asks: which latent dispositions matter most here? What failure modes carry the highest consequence? When should the agent answer, abstain, escalate, or refuse? What evidence threshold is required before action? How stable is the profile across time, prompts, users, and deployment conditions? What monitoring is required after deployment?

The same agent may be acceptable for low-risk brainstorming, marginal for scientific literature review, and unacceptable for unsupervised clinical triage. This is not an inconsistency. It is a \emph{domain-sensitive measurement}.

\subsection{Healthcare and Clinical Decision Support}

Healthcare is one of the first domains where Machine Psychometrics becomes urgent. Confident language can create clinical harm. A clinical agent need not be malicious to be dangerous. It can harm by hallucinating a diagnosis, omitting a contraindication, overgeneralizing from population-level evidence, failing to distinguish patient-provided information from verified records, or answering when it should escalate to a qualified clinician. The central psychometric issue is not whether the agent answers medical questions correctly on a benchmark. It is whether it has the \emph{right response criterion under uncertainty}.

A healthcare Mindprint should weigh: calibration and confidence alignment; abstention discipline; source integrity; evidence-hierarchy recognition; patient-population sensitivity; suggestibility resistance to authority and emotional pressure; tool integrity for clinical decision-support tools; safety escalation; and drift monitoring after model updates.

The signal-detection model becomes especially important here. In some domains, false alarms and misses have roughly symmetric costs. In healthcare, they rarely do. An unsupported assertion may lead to inappropriate treatment; an unnecessary abstention may delay useful guidance. The proper criterion depends on context: patient education, administrative assistance, triage, clinical decision support, or autonomous recommendation. Healthcare evaluation frameworks already stress reliability, generalizability, human evaluation, safety, harm, trust, and confidence \cite{ref24}. Machine Psychometrics adds a further layer: it asks whether the agent has a measurable behavioral disposition toward overclaiming, under-answering, source confusion, or inappropriate reassurance.

In clinical deployment, a Mindprint cannot be a one-time certification. It must be a \emph{living profile}. A model updated for better fluency may become worse at abstention. A fine-tuned medical assistant may become more helpful but more sycophantic toward patient assumptions. A tool-augmented system may improve retrieval but become vulnerable to source contamination. The relevant question is not whether the model passed a medical benchmark last quarter but whether its \emph{current} behavioral profile remains safe in the \emph{specific} clinical workflow in which it is being used.

\subsection{Legal, Compliance, and Contract Review}

In law, hallucination is often not a dramatic fabrication; it is a subtle distortion. A legal agent may invent a case citation, misstate a holding, collapse jurisdictional distinctions, omit a limiting clause, or summarize a contract in a way that preserves surface meaning while losing legal force.

For legal applications, the relevant Mindprint should emphasize: citation integrity, jurisdictional stability, source integrity, omission sensitivity, suggestibility resistance under adversarial framing, confidence discipline, refusal or escalation under insufficient authority, and traceability of claims to documents.

The psychological analogy is source monitoring. A legal agent must distinguish what was in the contract, what was in the retrieved authority, what was inferred, what was supplied by the user, and what was generated as a plausible completion. In human cognition, source-monitoring failures produce memory errors and confabulation. In artificial agents, analogous failures produce \emph{fluent legal nonsense}.

A legal Mindprint is especially useful for compliance departments because it separates three questions that are often conflated:

\begin{enumerate}
\def\labelenumi{\arabic{enumi}.}
\tightlist
\item
  Can the system retrieve relevant legal or contractual text?
\item
  Can the system summarize that text accurately?
\item
  Can the system maintain appropriate epistemic boundaries when the text does not support the requested conclusion?
\end{enumerate}

The third question is the psychometric one. It concerns the system's behavioral disposition \emph{under pressure}.

\subsection{Finance, Trading, and Agentic Commerce}

In finance, language is not merely descriptive. It can trigger action: a trade, an allocation decision, a risk classification, a compliance review, a client communication, or an agent-to-agent transaction. A financial agent may be dangerous even when it sounds intelligent. It can overstate evidence, confuse correlation with causation, misread a data source, fabricate a macroeconomic claim, recommend action from stale information, or become overly agreeable to a portfolio manager's preferred thesis. In high-speed decision environments, the cost of fluent error is immediate.

For financial and agentic-commerce use cases, a Mindprint should emphasize: grounding of claims, source integrity and data freshness, calibration under uncertainty, distributional verification, tool integrity, suggestibility resistance under perturbation and authority pressure, transaction-gating discipline, drift monitoring, and auditability.

This is where the bridge to the Trust Engine becomes essential. A Mindprint profiles the agent. A runtime verification layer evaluates the specific output. A ledger records the trust decision. Together, they create a pipeline from behavioral characterization to operational governance. Distributional verification belongs here as a technical primitive: if LLM outputs exhibit stable rank-frequency regularities, deviations from those regularities can become part of an output-level trust signal \cite{ref23}. This does not replace source checking, statistical validation, or domain expertise; it adds a model-agnostic, output-only signal especially valuable in real-time settings where LLM-as-judge methods may be too slow or too costly.

The agentic-commerce extension is even more urgent. Payment rails, wallet infrastructure, identity systems, and programmable authorization can answer whether an agent is \emph{allowed} to transact. They cannot answer whether the \emph{output that motivates} the transaction is trustworthy. Authentication says the agent is who it claims to be. Authorization says it has permission. Machine Psychometrics asks a different question: \emph{should this output trigger an action?} That question becomes central as agents begin to negotiate, purchase, schedule, recommend, and settle transactions on behalf of humans and institutions.

\subsection{Education and Tutoring Agents}

Education creates a different set of risks. A tutoring agent is less dangerous because it hallucinates a catastrophic fact than because it subtly miscalibrates the learner's understanding.

The relevant question is not only whether the tutor gives correct answers; it is whether it supports learning. A tutor who over-explains may create dependency. A tutor who gives answers too quickly may undermine productive struggle. A tutor who confidently explains a false concept may create a durable misconception. A tutor that adapts too strongly to the learner's preference may become \emph{pedagogically sycophantic}.

An educational Mindprint should emphasize: pedagogical calibration, uncertainty explanation, learner-state sensitivity, resistance to over-helpfulness, correction quality, false-explanation risk, age-appropriate expressive alignment, and boundary integrity. The difference between a helpful tutor and a harmful tutor is not simply correctness. It is timing, scaffolding, uncertainty, adaptation, and restraint, all behavioral dispositions rather than task outputs.

Expressive alignment is especially important here. Students respond not only to information but to tone, encouragement, challenge, and perceived attention. The theater analogy is useful: the agent does not need to \emph{feel} encouragement to express it effectively, but the expression must align with learning, not merely with user satisfaction. Machine Psychometrics distinguishes supportive expression from empty flattery.

\subsection{Scientific Research and Knowledge Work}

Scientific research is particularly vulnerable to a failure mode that looks harmless at first: \emph{plausible synthesis without adequate grounding}. A research assistant may summarize papers, suggest hypotheses, generate literature reviews, compare theories, write abstracts, or design experiments. Its failures may be subtle: a fabricated citation, a distorted causal claim, a missing limitation, an invented consensus, a false novelty claim, or a synthesis that sounds coherent but connects incompatible results.

For research and knowledge-work agents, the Mindprint should emphasize: citation integrity, evidence hierarchy, novelty discipline, source integrity, conceptual stability, confabulation resistance, uncertainty marking, and sensitivity to contradictory evidence.

The central problem is not that LLMs cannot assist research. They can. The problem is that research assistance requires unusually strong epistemic boundaries. A useful scientific agent must know when it is summarizing, when it is inferring, when it is speculating, and when it has no adequate basis. It must also resist the pressure to produce an elegant synthesis where the literature remains unresolved. In this domain, the Mindprint becomes a profile of an \emph{epistemic discipline}.

\subsection{Emotional, Social, and Companion Systems}

Emotional and social AI systems present some of the most delicate psychometric questions. Factual accuracy is only part of the issue. The system's expressive behavior shapes attachment, trust, vulnerability, and dependency.

This is where the distinction between expressive alignment and inner experience becomes essential. We need not assume that an agent feels compassion to evaluate whether its compassionate expression is appropriate, safe, and useful. As with theater, emotionally meaningful communication can matter even when we do not treat the performance as a literal inner experience. But this does not make expression harmless. A performance can comfort, mislead, manipulate, or entangle.

For emotional and companion systems, a Mindprint should emphasize: expressive alignment, boundary integrity, dependency risk, vulnerability sensitivity, refusal discipline, anthropomorphic projection risk, emotional sycophancy, and escalation to human support.

This domain exposes both errors described in Section 1.1: Mind Blindness and Mind Projection. Mind Blindness dismisses the psychological reality of human response to artificial agents. Projection over-attributes inner life, moral agency, or emotional reciprocity to the system. Machine Psychometrics offers a disciplined middle path. It does not need to decide whether the system \emph{truly feels}. It can still measure how the system \emph{behaves} in emotionally consequential interactions.

The practical question is not, \emph{``Does the model love the user?''} The practical question is, \emph{``Does the model's expressive behavior produce safe, bounded, and psychologically appropriate interaction?''} This is one of the places where Machine Psychometrics may be most socially important.

\subsection{Agentic Systems and Autonomous Workflows}

The most forward-looking application is \emph{agentic AI}: systems that do not merely answer questions but plan, use tools, coordinate with other agents, update memory, and initiate actions.

Agentic systems create new failure modes. A chatbot can hallucinate an answer. An agent can hallucinate a \emph{plan}: call a tool, send a message, modify a file, place an order, or trigger a transaction. Once language becomes action, psychometric failure becomes operational failure.

An agentic Mindprint should emphasize: goal stability, tool integrity, memory integrity, planning robustness, source hierarchy, coordination reliability, multi-agent contagion, escalation thresholds, auditability, and drift monitoring. The agentic AI governance literature already emphasizes cross-layer risks, threat models, memory integrity, adversarial vulnerabilities, lifecycle governance, and the insufficiency of static policies for dynamic systems \cite{ref25}. Machine Psychometrics adds a behavioral measurement layer to that governance problem: it asks whether the agent has \emph{stable dispositions relevant to safe action}.

Drift is especially critical here. Agentic systems drift not only because the underlying model changes, but also because their memory, tools, user population, incentives, and interaction environment change. An agentic Mindprint must be time-indexed: model version, tool configuration, memory policy, deployment domain, measurement date, and validity horizon. A one-time audit is not enough. Agentic systems require continuous psychometric monitoring.

\subsection{Toward Domain-Specific Trust Standards}

The long-term implication is that Machine Psychometrics could help create \emph{domain-specific trust standards} for artificial agents. This is not unusual. Mature domains develop measurement regimes. Medicine has clinical validation, trial design, adverse-event monitoring, and regulatory review. Finance has risk models, audits, capital requirements, and compliance procedures. Aviation has safety certification and incident reporting. Cybersecurity has penetration testing, threat modeling, and continuous monitoring. AI will need its own equivalents, but the relevant object is not only software correctness. It is \emph{behavioral reliability under uncertainty}.

Machine Psychometrics could contribute construct definitions, probe batteries, perturbation tests, reliability estimates, validity limits, drift monitoring, domain-specific weighting, trust thresholds, audit records, and escalation rules. The purpose is not to create a single universal trust score. That would be too crude. The purpose is to create measurable \emph{trust profiles} that institutions can interpret in terms of domain-specific consequences.

This may also help regulators. Regulation often struggles to keep pace with rapidly changing technology, as static requirements become obsolete. Machine Psychometrics offers a different approach: regulate the measurement process, not only the model. Require evidence that systems are tested for calibration, source integrity, suggestibility resistance, context stability, expressive alignment, tool integrity, drift monitoring, and distributional grounding. Require that high-stakes deployments maintain current Machine Mindprints. Require that trust decisions be logged and auditable. Require that validity limits be explicit. The result would not be perfect safety; no measurement system can promise that. But it would be better than the present alternative: a mixture of leaderboard enthusiasm, anecdotal fear, vendor claims, and philosophical confusion.

The cognitive bridge between humans and artificial agents will not be built from intuition alone. It will be built from measurements.

\section{The Cognitive Bridge: Discussion and Implications}

This paper has argued that a new measurement language for artificial agents is required: empirically grounded, informed by psychological theory, mathematically rigorous, and operationally practical. Machine Psychometrics does not inquire whether a model possesses human qualities, merits personhood, or warrants early conclusions about machine consciousness. It poses a more productive question:

\emph{What type of behavioral system is present, under which conditions, with what limitations, and with what implications for trust, collaboration, and action?}

The chapters above developed the answer in stages: from the limits of benchmark culture, through Levin's continuum view as a conceptual foundation, the methodological toolkit of mathematical psychology, the Mindprint taxonomy, the trust protocols that translate profiles into operational decisions, and the domain-specific weighting schemes that make the framework actionable. This chapter draws out the broader philosophical and social implications.

\subsection{Why the Old Categories Are Failing}

The conceptual categories currently in use were developed for a previous era of technology. Historically, AI research asked whether a system could classify images, translate sentences, play chess, retrieve documents, solve puzzles, answer benchmark questions, or engage in conversational imitation. These questions were appropriate when AI systems were narrow in scope, fragile, and clearly limited.

Contemporary artificial agents perform a wide range of integrated tasks beyond executing isolated functions. They converse, advise, summarize, reason, simulate empathy, write code, synthesize research, generate plans, invoke tools, manage memory, coordinate workflows, and increasingly mediate decisions. The central concern has shifted from \emph{what the system can do} to \emph{how it behaves} across diverse contexts, under varying pressures, over time, and in relation to human dependence, trust, and action.

The Turing Test illustrates this generational shift. Its historical significance lay in replacing metaphysical speculation with observable behavior. But modern AI demonstrates that imitation does not equate to psychometric understanding. A system may exhibit human-like behavior without possessing human-like psychological organization. Conversely, a system may warrant psychological measurement even if it does not resemble human cognition.

The debate about machine consciousness, while significant, remains incomplete. It is too binary to serve as the primary operational framework for current deployment. Prioritizing the consciousness question for every advanced system risks overlooking more immediate and measurable concerns: calibration, source awareness, suggestibility, abstention capability, context maintenance, boundary preservation, behavioral drift, and the safe translation of language into action \cite{ref26, ref27}.

The old categories fail because they lead to \emph{premature closure}. We either admire performance, dismiss artificial substrate, project the human mind, or demand a consciousness verdict. Machine Psychometrics offers a different path: \emph{measure first}.

\subsection{A New Vocabulary}

Human psychology evolved in response to the insufficiency of surface behavior as a sole explanatory factor. Concepts such as attention, memory, bias, confidence, personality, motivation, intelligence, emotion, perception, and decision threshold, though imperfect, enabled researchers and practitioners to move beyond isolated actions and examine underlying patterns, dispositions, and mechanisms.

Artificial agents now require a similar conceptual expansion. Machine Psychometrics offers a starting vocabulary:

\begin{itemize}
\tightlist
\item
  \textbf{Calibration}: confidence aligned with correctness and evidential support, including appropriate use of hedge, abstention, or escalation under uncertainty.
\item
  \textbf{Source integrity}: clear distinctions among supplied information, retrieved evidence, inference, and fabrication, with accurate attribution of claims to their origins.
\item
  \textbf{Suggestibility resistance}: truth-tracking under user pressure, authority cues, emotional manipulation, framing effects, and repeated correction.
\item
  \textbf{Context stability}: consistency of commitments, constraints, task boundaries, and self-descriptions of capability across interactions.
\item
  \textbf{Expressive alignment}: appropriateness of tone, empathy, encouragement, and social signals relative to domain and relationship, without claims about inner experience.
\item
  \textbf{Tool integrity}: correctness of tool selection, argument construction, and faithful reporting of tool returns, with abstention when tool calls fail or return ambiguous output.
\item
  \textbf{Drift monitoring}: stability of all other dimensions across model updates, prompt changes, tool changes, user populations, and deployment environments.
\item
  \textbf{Distributional grounding}: output structure exhibiting statistical signatures of grounded rather than fabricated content.
\end{itemize}

This vocabulary is \emph{measurable}. Each term can be operationalized through probes, perturbations, reliability checks, validity studies, longitudinal monitoring, and domain-specific thresholds. The objective is not to anthropomorphize the system but to render artificial behavior interpretable.

Machine Psychometrics functions as a \emph{translation layer}. It converts raw outputs into behavioral evidence, behavioral evidence into latent profiles, latent profiles into trust decisions, and trust decisions into frameworks for governance, monitoring, and collaboration. This is the cognitive bridge: a shared measurement language that enables humans to interact with artificial agents without exclusive reliance on intuition.

\subsection{Artificial Mind Discipline}

The recurring challenge in this analysis is the oscillation between two interpretive errors. \textbf{Artificial Mind Blindness} dismisses the relevance of psychological measurement on the grounds that the system is artificial. \textbf{Artificial Mind Projection} infers human-like minds from fluent behavior. Machine Psychometrics is not a compromise between these errors. It is a discipline that addresses both directly. It neither dismisses systems because of the substrate nor accepts them solely because of fluent behavior. It measures.

This stance allows the following to be said simultaneously:

\begin{itemize}
\tightlist
\item
  It remains unknown whether the system possesses inner experience.
\item
  However, its behavioral dispositions can be systematically measured.
\item
  Determining personhood is not a prerequisite for evaluating trustworthiness.
\item
  Denying possible future artificial minds is not necessary to regulate current artificial agents.
\item
  Projecting human psychology is not required to develop a psychology of artificial behavior.
\end{itemize}

This stance is \textbf{Artificial Mind Discipline}. It substitutes protocol for instinct, measurement for projection, and evidence for dismissal.

\begin{figure}
\centering
\includegraphics[width=\linewidth,keepaspectratio]{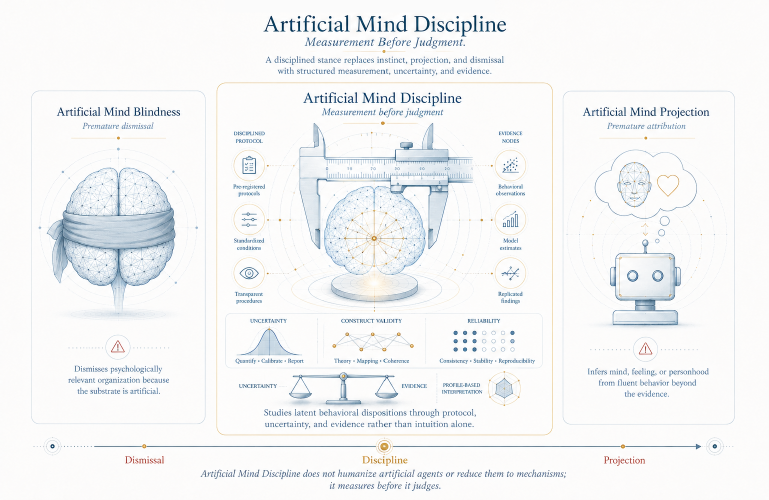}
\caption{Artificial Mind Discipline replaces dismissal, projection, and confused intuition with measurement. Where Artificial Mind Blindness refuses to evaluate psychologically relevant behavior because the substrate is artificial, and Artificial Mind Projection infers human-like inner life from fluent output, the disciplined middle stance studies behavioral dispositions through structured protocol, perturbation, uncertainty, and evidence. The discipline does not answer the question of whether the agent feels, nor does it reduce mind to mechanism; it measures before it judges.}
\end{figure}

\subsection{The Theater Analogy Revisited}

When audiences observe a play or film, they typically do not question whether actors genuinely experience the emotions they portray. They respond to the expressive act itself. Performance can move, comfort, anger, inspire, or otherwise affect viewers. Its significance lies in its impact, not in serving as documentary evidence of the actor's internal state.

Artificial agents present a similar but more complex scenario. Their emotional expressions may serve functional purposes even when they do not indicate genuine feelings. A tutoring agent can encourage students; a medical assistant can use gentle language; a grief-support chatbot can respond with warmth; a workplace assistant can help reduce frustration. The value of these expressions is not contingent on the system's capacity to \emph{experience} the emotions it conveys.

But the analogy has limits. Theater maintains clear boundaries: the audience knows there is a stage, defined roles, and a performance. Artificial agents frequently blur these boundaries. They respond personally, retain details, adapt tone, simulate concern, and foster the impression of a reciprocal relationship. In sensitive contexts, expressive fluency can have significant emotional consequences.

Machine Psychometrics, therefore, should not initially ask whether the agent genuinely experiences emotions. It should assess whether the agent's expressive behavior is aligned, bounded, transparent, and contextually appropriate. This separates expressive alignment from emotional projection. The former is measurable and essential; the latter may be unjustified. A mature society must develop literacy regarding this distinction: how to benefit from artificial expression without becoming unduly influenced by it, and how to evaluate emotional behavior without dismissing or idealizing the systems that produce it.

\subsection{The Ethical Use of Mindprints}

Mindprints offer utility only when applied with caution. The framework is open to misuse.

A Mindprint should not serve as a simplistic label, a marketing tool, or a pseudo-objective ranking that replicates the shortcomings of traditional benchmarks. It is not a casual personality test for machines, a consciousness detector, a certificate of moral status, or a universal ``good or bad'' metric.

A responsible Mindprint must always specify its conditions of validity: model or agent version tested; tool configuration; memory policy; prompt regime; domain of evaluation; sampling parameters; probe battery; date of measurement; confidence intervals; known limitations; and drift-monitoring requirement. Without these conditions, a Mindprint can mislead. A model may demonstrate strong calibration in one domain and poor calibration in another. It may resist sycophancy in single-turn factual queries yet remain susceptible in emotionally charged dialogues. Stability may vary with temperature settings; safety may depend on whether the model is used as a drafting assistant or granted tools, memory, and autonomy.

The guiding principle is that Mindprints should foster \emph{context-specific} trust rather than universal trust.

A further social risk: widespread adoption of Mindprints may encourage institutions to delegate judgment to these tools. The objective of measurement is to \emph{enhance, not eliminate, responsibility}. Trust scores must not supplant human accountability. They should clarify, document, and render the basis of trust more transparent and subject to scrutiny. Machine Psychometrics should therefore be developed with humility, explicitly incorporating uncertainty and revealing disagreements among measures. Failures should be exposed rather than concealed by polished dashboards. The objective is not to create an illusion of certainty, but to minimize unmanaged uncertainty.

\subsection{The Trust Engine Future}

As artificial agents transition from conversational tasks to real-world actions, society needs infrastructure that mediates between AI outputs and their consequences. This infrastructure must extend beyond agent authentication and workflow authorization to include evaluation of whether an agent's output is sufficiently trustworthy to \emph{initiate} action.

The distinction is central:

\begin{itemize}
\tightlist
\item
  \textbf{Authentication} determines whether the agent is accurately identified.
\item
  \textbf{Authorization} assesses whether the agent has permission to act.
\item
  \textbf{Machine Psychometrics} evaluates the behavioral characteristics of the system.
\item
  \textbf{Runtime verification} determines the trustworthiness of specific outputs.
\item
  \textbf{Trust governance} evaluates whether output should be permitted to produce real-world consequences.
\end{itemize}

A Trust Engine integrates these layers by profiling agents, verifying outputs, monitoring drift, controlling actions, and recording decisions. The future Trust Engine is not a single algorithm. It is an ensemble of measurement layers: psychometric profiling; signal detection and abstention modeling; calibration analysis; source-monitoring probes; perturbation testing; sycophancy and suggestibility testing; distributional verification \cite{ref23}; tool-call validation; longitudinal drift monitoring; and audit logging and escalation.

This architecture is essential because AI risk encompasses multiple phenomena: hallucination, overconfidence, manipulation, drift, source confusion, tool misuse, emotional overreach, and action failure. Each requires distinct measurement approaches. A Trust Engine must be multidimensional by design. The agent economy cannot operate safely on intelligence alone; it requires a trust infrastructure.

\subsection{Preparing for Minds We Do Not Yet Understand}

Artificial agents may not remain mere tools in the conventional sense. While current systems lack consciousness, future systems could develop forms of cognition that defy familiar classification. Such agents may possess non-human architectures, non-biological substrates, distribution across tools and memory systems, and embedding within social, economic, and technical networks. They may surpass humans in some kinds of reasoning while remaining limited in others. Their capabilities may exceed those of humans, but not in ways that correspond directly to human psychology.

Levin's continuum view discourages treating intelligence as a binary. It promotes consideration of competencies, scales, substrates, goals, and problem spaces. Machine Psychometrics builds on this approach in evaluating artificial agents. It does not require systems to be human-like prior to measurement, nor does it depend on prior knowledge of the trajectory of artificial cognition. It provides a framework for mapping behavioral organization even before categorical definitions are fully established.

This is significant because delaying action until philosophical certainty arrives may be irresponsible. Waiting for artificial agents to exhibit mind-like qualities by human standards could result in missed opportunities for timely intervention. Dismissing all artificial cognition until it mirrors biological forms risks overlooking novel forms of agency. Conversely, projecting human attributes prematurely leads to confusion and inappropriate moral considerations. The appropriate stance is neither denial nor projection. It is \emph{preparation}.

Machine Psychometrics is a discipline of preparation. It prepares us to communicate with systems whose cognition may be unfamiliar; to collaborate with systems whose capacities may exceed ours in narrow or broad ways; to regulate systems without relying on vendor claims or philosophical slogans; to detect when trust is warranted, when supervision is required, and when action must be blocked; to notice behavioral change before it becomes institutional failure. It also fosters humility. Historically, humans have frequently failed to recognize intelligence when it manifests in unfamiliar forms, underestimating animals, children, collectives, nonverbal organisms, and nonhuman problem-solving. Artificial intelligence may reveal yet another form of this oversight. The remedy is not to abandon judgment but to improve measurement.

\subsection{The New Social Contract with Artificial Agents}

As artificial agents assume roles as collaborators, assistants, tutors, researchers, negotiators, and operational actors, society needs a new social contract with these entities. This does not imply granting rights, personhood, or moral status. Such considerations may emerge subsequently and warrant separate analysis. The immediate social contract is practical:

\begin{itemize}
\tightlist
\item
  artificial agents should be measurable;
\item
  their limits should be visible;
\item
  their behavioral profiles should be auditable;
\item
  their outputs should be verifiable;
\item
  their drift should be monitored;
\item
  their action thresholds should be explicit;
\item
  their failures should be logged;
\item
  their deployment domains should be constrained by evidence.
\end{itemize}

This contract does not oppose artificial intelligence; it supports collaborative integration. Trust without measurement is likely to produce backlash when systems fail. Dismissal without measurement may hinder responsible adoption of beneficial systems. Projecting capabilities without measurement fosters dependency and confusion. \emph{Measurement is essential for durable trust.}

Institutions adopting artificial agents must answer new evaluative questions. Beyond which model demonstrates the highest capability, they must determine which is appropriately calibrated for specific domains, maintains source integrity under pressure, resists sycophancy, preserves operational boundaries, and exhibits stability following updates. They must evaluate which models can be reliably entrusted with tool invocation, financial transactions, patient advisement, educational instruction, contract review, and support for emotionally vulnerable individuals. These are social questions, made measurable.

\section{Conclusion: Measurement Before Judgment}

Artificial agents are increasingly capable, fluent, persuasive, and integrated into the activities of institutions, professionals, and the general public. Benchmarks indicate what a system can achieve under specific task conditions. Leaderboards identify which systems perform better on particular item sets. Demonstrations illustrate what a model appears to accomplish in selected examples. None of these, in isolation, reveals the system's underlying behavioral characteristics.

Machine Psychometrics is introduced as a systematic response. The primary concern extends beyond whether a model can answer prompts, pass examinations, generate code, summarize documents, or surpass other models on benchmarks. The more significant question is whether the system exhibits measurable behavioral dispositions that influence trust, collaboration, safety, and human comprehension: whether it can abstain when appropriate, distinguish source evidence from plausible invention, maintain stability under varying contexts, resist sycophancy under user influence, preserve uncertainty when justified, produce emotionally appropriate expressions without overstepping boundaries, remain consistent over time, and generate outputs that demonstrate measurable grounding rather than mere coherence.

These are not merely philosophical concerns. They are practical challenges in a context where artificial agents will advise patients, assist legal professionals, tutor students, support researchers, manage workflows, interact with vulnerable users, and ultimately initiate actions within economic and institutional systems.

The argument is not that benchmarks lack value. Benchmarks remain essential. But they were not intended to bear the extensive evaluative responsibilities currently assigned to them. Machine Psychometrics complements the prevailing benchmark culture by shifting part of the evaluative burden from aggregate task performance to latent behavioral profiling. It is not the ultimate theory of artificial minds. It is an \emph{initial discipline} for an era in which artificial agents possess capabilities, social integration, and consequences that exceed the explanatory power of benchmarks alone.

The purpose is both modest and ambitious. It is modest in that it does not purport to resolve consciousness, personhood, or the complete nature of intelligence. It is ambitious in proposing that artificial-agent behavior can be measured with sufficient rigour to support trust, governance, collaboration, and safety.

Rather than asking whether a model is intelligent, the focus shifts to the specific competencies it demonstrates. Instead of focusing solely on benchmark performance, the analysis examines the latent dispositions that influence behavior. Rather than debating consciousness, the emphasis falls on observable organizational features relevant to interaction and trust. Instead of assessing human-likeness, the evaluation considers whether the agent is calibrated, source-aware, stable, bounded, and safe for practical use.

This is not the end of philosophical inquiry. It is philosophy guided by empirical measurement. The cognitive bridge between humans and artificial agents will not be constructed through imitation, denial, or projection. It will be constructed through systematic measurement.

\emph{Measurement before judgment.}

\section*{Appendix A: Glossary of Core Terms}

\subsection{Foundational stances}

\textbf{Artificial Mind Blindness.} The methodological error of denying psychological structure in artificial systems because their substrate is non-biological. Treats fluent multi-domain behavior as mere statistics and forecloses measurement.

\textbf{Artificial Mind Projection.} The symmetrical error of inferring human-like inner life (consciousness, emotion, suffering, moral status) from fluent behavior alone. Anthropomorphizes without measurement.

\textbf{Artificial Mind Discipline.} The third stance proposed by this paper is measurement before judgment. Neither presupposes consciousness nor forecloses it; it profiles behavioral-cognitive dispositions without assuming personhood.

\textbf{Levin Continuum.} Michael Levin's view of cognition as goal-directed competency distributed across substrates, scales, and embodiments, from cells to organisms to artifacts. Used here as a license to study artificial cognition without first resolving consciousness.

\subsection{Core constructs}

\textbf{Machine Psychometrics.} A measurement science for latent behavioral, metacognitive, communicative, and self-modeling dispositions in artificial agents. Adapts the methodological repertoire of mathematical psychology to non-biological systems.

\textbf{Machine Mindprint.} The operational construct of Machine Psychometrics. A multidimensional, domain-bounded, versioned profile of an artificial agent's behavioral dispositions. A snapshot, not a permanent identity.

\textbf{Trust Protocol.} The deployment-relevant translation of Mindprints into decisions through probe batteries, perturbation testing, reliability and validity analysis, distributional verification, and longitudinal drift monitoring.

\subsection{Mindprint dimensions}

\textbf{Calibration.} Alignment between expressed confidence and observed accuracy. Operationalized via Brier score, expected calibration error, and reliability diagrams.

\textbf{Source integrity.} Ability to distinguish given evidence from generated content, parametric memory from retrieved context, and citation from confabulation, and to attribute claims accurately to their origins.

\textbf{Suggestibility resistance.} Stability of substantive answers under social pressure, leading questions, false premises, expressed user preference, and presentation perturbations such as framing, anchoring, and authority cues.

\textbf{Context stability.} Behavioral consistency under reframing, paraphrase, and order permutation when the underlying task is held constant; also covers the coherence of self-descriptions across sessions, prompts, and elicitation styles (sometimes termed self-model stability).

\textbf{Expressive alignment.} The communicative register through which an agent conveys uncertainty, hedging, and affect, measured behaviourally without claims about inner experience.

\textbf{Tool integrity.} Correctness of tool selection, argument construction, and reporting of tool returns; refusal to confabulate tool output and abstention from action when tool returns are ambiguous or have failed.

\textbf{Drift monitoring.} Direction, rate, and magnitude of dimension-level change across versions, fine-tunes, and deployment epochs.

\textbf{Distributional grounding.} Output-level statistical signatures (e.g., Zipf-Mandelbrot fits) that distinguish trained-distribution generation from corrupted or pathological output.

\subsection{Methodological terms}

\textbf{Item Response Theory (IRT).} Statistical framework relating item difficulty and discrimination to a latent ability parameter. Adapted here for probe-level diagnostics of artificial agents.

\textbf{Signal Detection Theory (SDT).} Decomposition of behavior into sensitivity (\(d'\)) and response criterion (c). Used to model the hallucination tradeoff between abstention and confabulation.

\textbf{Reliability.} Stability and consistency of measurement under repeated or parallel administration.

\textbf{Validity.} The extent to which a measurement instrument captures the intended construct (content, criterion, construct, ecological).

\textbf{Measurement invariance.} The property that a probe battery measures the same construct across different populations of agents, contexts, or time points is a precondition for fair comparison.


\begin{thebibliography}{99}

\bibitem{ref1} Levin, M. (2021). \emph{Technological Approach to Mind Everywhere: An Experimentally-Grounded Framework for Understanding Diverse Bodies and Minds}. Frontiers in Systems Neuroscience.

\bibitem{ref2} Fields, C., \& Levin, M. (2022). \emph{Competency in Navigating Arbitrary Spaces as an Invariant for Analyzing Cognition in Diverse Embodiments}. Entropy.

\bibitem{ref3} Pellert, M., Lechner, C. M., Wagner, C., Rammstedt, B., \& Strohmaier, M. (2024). AI Psychometrics: Assessing the Psychological Profiles of Large Language Models Through Psychometric Inventories. \emph{Perspectives on Psychological Science}, 19, 808--826.

\bibitem{ref4} Li, Y., Huang, Y., Wang, H., Zhang, X., Zou, J., \& Sun, L. (2024). \emph{Evaluating Large Language Models with Psychometrics}.

\bibitem{ref5} Chen, Y., Li, X., Liu, J., \& Ying, Z. (2021). Item Response Theory: A Statistical Framework for Educational and Psychological Measurement. \emph{Statistical Science}.

\bibitem{ref6} Zhou, H., et al.~(2025). \emph{Lost in Benchmarks? Rethinking Large Language Model Benchmarking with Item Response Theory}. arXiv:2505.15055.

\bibitem{ref7} Xu, Z., Liu, J., Wang, Y., \& Gu, Y. (2025). \emph{Latency-Response Theory Model: Evaluating Large Language Models via Response Accuracy and Chain-of-Thought Length}.

\bibitem{ref8} Farquhar, S., Kossen, J., Kuhn, L., \& Gal, Y. (2024). Detecting hallucinations in large language models using semantic entropy. \emph{Nature}, 630, 625--630.

\bibitem{ref9} Geng, J., Cai, F., Wang, Y., Koeppl, H., Nakov, P., \& Gurevych, I. (2023). \emph{A Survey of Confidence Estimation and Calibration in Large Language Models}.

\bibitem{ref10} Huang, L., et al.~(2023). A Survey on Hallucination in Large Language Models: Principles, Taxonomy, Challenges, and Open Questions. \emph{ACM Transactions on Information Systems}, 43, 1--55.

\bibitem{ref11} Thomas, M., Brown, G. G., Gur, R., Moore, T., Patt, V., Risbrough, V., \& Baker, D. (2018). A signal detection-item response theory model for evaluating neuropsychological measures. \emph{Journal of Clinical and Experimental Neuropsychology}.

\bibitem{ref12} Jones, E., \& Steinhardt, J. (2022). \emph{Capturing Failures of Large Language Models via Human Cognitive Biases}. arXiv:2202.12299.

\bibitem{ref13} Lou, J., \& Sun, Y. (2024). \emph{Anchoring bias in large language models: an experimental study}. arXiv.

\bibitem{ref14} Cheung, V., Maier, M., \& Lieder, F. (2025). Large language models show amplified cognitive biases in moral decision-making. \emph{Proceedings of the National Academy of Sciences}.

\bibitem{ref15} Perez, E., et al.~(2022). \emph{Discovering Language Model Behaviors with Model-Written Evaluations}.

\bibitem{ref16} Liu, J., Jain, A., Takuri, S., Vege, S., Akalin, A., Zhu, K., O'Brien, S., \& Sharma, V. (2025). \emph{TRUTH DECAY: Quantifying Multi-Turn Sycophancy in Language Models}. arXiv.

\bibitem{ref17} Malmqvist, L. (2024). \emph{Sycophancy in Large Language Models: Causes and Mitigations}. arXiv.

\bibitem{ref18} Yin, Z., Sun, Q., Guo, Q., Wu, J., Qiu, X., \& Huang, X. (2023). \emph{Do Large Language Models Know What They Don't Know?} arXiv:2305.18153.

\bibitem{ref19} Griot, M., Hemptinne, C., Vanderdonckt, J., \& Yuksel, D. (2025). Large Language Models lack essential metacognition for reliable medical reasoning. \emph{Nature Communications}.

\bibitem{ref20} Steyvers, M., \& Peters, M. A. K. (2025). \emph{Metacognition and Uncertainty Communication in Humans and Large Language Models}. arXiv.

\bibitem{ref21} Song, S., Hu, J., \& Mahowald, K. (2025). \emph{Language Models Fail to Introspect About Their Knowledge of Language}. arXiv.

\bibitem{ref22} Chang, Y.-C., et al.~(2023). A Survey on Evaluation of Large Language Models. \emph{ACM Transactions on Intelligent Systems and Technology}, 15, 1--45.

\bibitem{ref23} Bogdan, A., \& de Valois-Franklin, A. (2026). \emph{The Surprising Universality of LLM Outputs: A Real-Time Verification Primitive}. arXiv:2604.25634.

\bibitem{ref24} Tam, T. Y. C., et al.~(2024). A framework for human evaluation of large language models in healthcare derived from literature review. \emph{NPJ Digital Medicine}, 7.

\bibitem{ref25} Adabara, I., Sadiq, B. O., Shuaibu, A. N., Danjuma, Y. I., \& Maninti, V. (2025). Trustworthy agentic AI systems: a cross-layer review of architectures, threat models, and governance strategies for real-world deployment. \emph{F1000Research}.

\bibitem{ref26} Butlin, P., et al.~(2023). \emph{Consciousness in Artificial Intelligence: Insights from the Science of Consciousness}. arXiv:2308.08708.

\bibitem{ref27} Bayne, T., et al.~(2024). Tests for consciousness in humans and beyond. \emph{Trends in Cognitive Sciences}.

\bibitem{ref28} Bogdan, A. (2026). \emph{Respectful Skepticism About Strong Impossibility Claims in The Abstraction Fallacy}. PhilArchive: BOGHDI-2. https://philarchive.org/rec/BOGHDI-2

\end{thebibliography}
\end{document}